\documentclass{article} 
\usepackage[dvipsnames]{xcolor}
\usepackage{iclr2021_conference,times}

\usepackage{amsmath,amsfonts,bm}

\def\eqref#1{equation~\ref{#1}}

\def\1{\bm{1}}

\DeclareMathAlphabet{\mathsfit}{\encodingdefault}{\sfdefault}{m}{sl}
\SetMathAlphabet{\mathsfit}{bold}{\encodingdefault}{\sfdefault}{bx}{n}

\usepackage{hyperref}
\usepackage{url}
\usepackage{graphicx}
\usepackage{multirow}
\usepackage[export]{adjustbox}
\usepackage{subcaption}
\usepackage{caption}
\usepackage{amssymb}
\usepackage{amsmath}
\usepackage{algorithm,algpseudocode}
\usepackage{longtable}
\usepackage{pbox}
\usepackage{booktabs}       
\usepackage{xspace}
\usepackage{wrapfig}
\usepackage{wasysym}
\usepackage{pifont}

\makeatletter
\DeclareRobustCommand\onedot{\futurelet\@let@token\@onedot}
\def\@onedot{\ifx\@let@token.\else.\null\fi\xspace}
\def\eg{\emph{e.g}\onedot} 
\def\ie{\emph{i.e}\onedot} 
 
\def\etc{\emph{etc}\onedot}

\makeatother

\title{Searching towards Class-Aware Generators for\!\!\!\!\!\!\!\!\!\!\!\!\\Conditional Generative Adversarial Networks\!\!\!\!\!\!}

\author{%
Peng Zhou$^1$, Lingxi Xie$^2$, Xiaopeng Zhang$^2$, Bingbing Ni$^1$, Qi Tian$^2$\\
$^1$Shanghai Jiao Tong University,\quad$^2$Huawei Inc.\\
{\tt\small zhoupengcv@sjtu.edu.cn}, {\tt\small 198808xc@gmail.com}, {\tt\small zxphistory@gmail.com},\\ {\tt\small nibingbing@sjtu.edu.cn}, {\tt\small tian.qi1@huawei.com}
}

\iclrfinalcopy 
\begin{document}

\maketitle

\begin{abstract}
   Conditional Generative Adversarial Networks (cGAN) were designed to generate images based on the provided conditions, \eg, class-level distributions. However, existing methods have used the same generating architecture for all classes. This paper presents a novel idea that adopts NAS to find a distinct architecture for each class. The search space contains regular and class-modulated convolutions, where the latter is designed to introduce class-specific information while avoiding the reduction of training data for each class generator. The search algorithm follows a weight-sharing pipeline with mixed-architecture optimization so that the search cost does not grow with the number of classes. To learn the sampling policy, a Markov decision process is embedded into the search algorithm and a moving average is applied for better stability. We evaluate our approach on CIFAR10 and CIFAR100. Besides achieving better image generation quality in terms of FID scores, we discover several insights that are helpful in designing cGAN models.
\end{abstract}

\section{Introduction}

Generative Adversarial Network (GAN)~\citep{goodfellow2014Generative} has attracted considerable attention and achieved great success in image generation.
Conditional GAN (cGAN)~\citep{mirza2014Conditional} is a type of GAN using class information to guide the training of the discriminator and generator so that it usually obtains a better generation effect. Most cGANs incorporate class information into the generator through Conditional Batch Normalization (\textit{CBN})~\citep{devries2017Modulating}, or into the discriminator through projection discriminator~\citep{miyato2018cGANs}, multi-hinge loss~\citep{kavalerov2019cGANs}, auxiliary loss~\citep{odena2017Conditional}, \etc.

In this paper, we investigate the possibility of designing class-aware generators for cGAN (\ie, using a distinct generator network architecture for each class). To automatically design class-aware generators, we propose a neural architecture search (NAS) algorithm on top of reinforcement learning so that the generator architecture of each class is automatically designed. However, as the number of classes increases, there are three main issues we have to consider. First, the search space will grow exponentially as the number of categories grows (\ie, combinatorial explosion). Second, training the generator separately for each class is prone to insufficient data~\citep{karras2020Training}. Furthermore, searching and re-training each generator one by one may be impractical when the number of generators is large.




We propose solutions for these challenges. First, we present a carefully designed search space that is both flexible and safe. We refer to \textbf{flexibility} as the ability to assign a distinct generator architecture to each class, which makes the search space exponentially large while its size is still controllable. To guarantee the \textbf{safety} (\ie, enable the limited amount of training data to be shared among a large number of generators), we introduce a new operator named Class-Modulated convolution (\textit{CMconv}). \textit{CMconv} shares the same set of convolutional weights with a regular convolution but is equipped with a standalone set of weights to modulate the convolutional weights, allowing the training data to be shared among different architectures and thus alleviating the inefficiency on training data. Second, to make the procedure of search and re-training as simple as possible, we develop \textit{mixed-architecture optimization}, such that the training procedure of multiple class-aware generators is as simple as that of training only one generator.

\begin{figure}[!t]
   \begin{center}
      \includegraphics[width=0.75\linewidth]{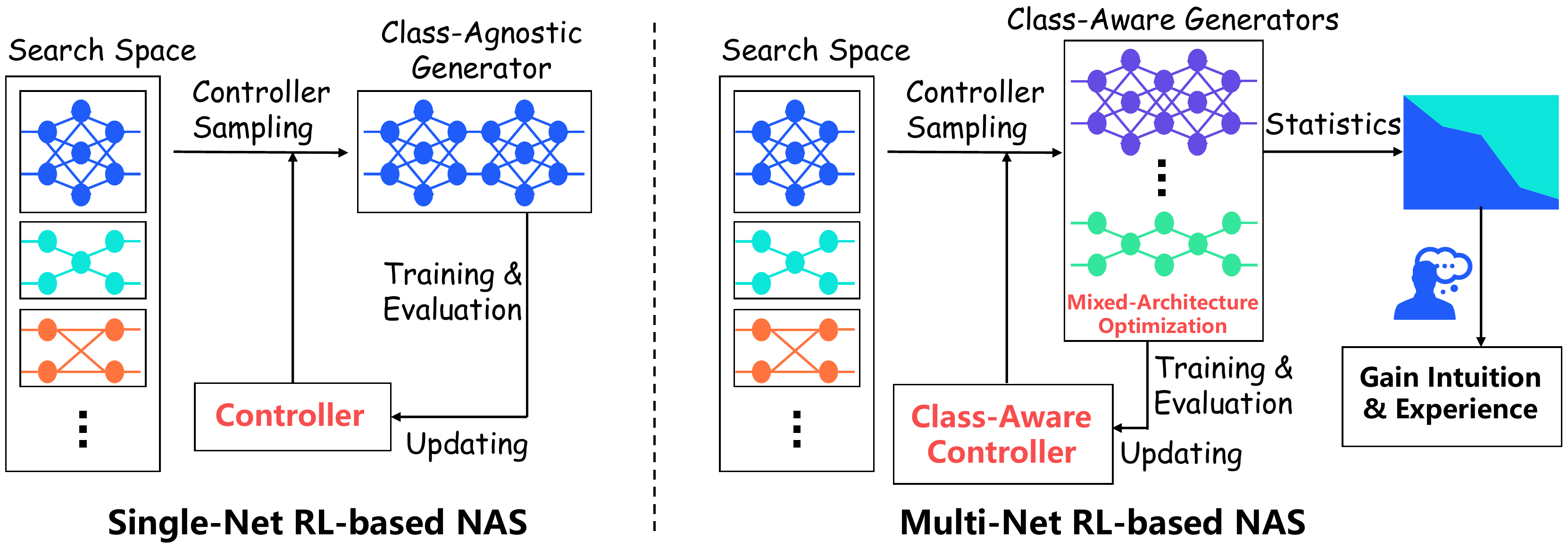}
   \end{center}
   \vspace{-0.4cm}
   \caption{Illustration of Single-Net RL-based NAS and our Multi-Net RL-based NAS. Class-aware generators allow us to discover statistical laws by analyzing multiple network architectures. However, this cannot be achieved by class-agnostic generator because only one generator is searched (\ie, one sample), so we cannot get more information for architecture design.}
   \label{fig:nas_framework}
\end{figure}

Integrating these modules produces the proposed Multi-Net NAS (MN-NAS). To the best of our knowledge, this is the first method that can produce a number of generator architectures, one for each class, through one search procedure. Figure~\ref{fig:nas_framework} shows the overall framework of MN-NAS. It applies a Markov decision process equipped with moving average as the top-level logic for sampling and evaluating candidate architectures. After the search procedure, the optimal architecture for each class is determined and they get re-trained and calibrated for better image generation performance.


We perform experiments on some popular benchmarks, including the CIFAR10 and CIFAR100 datasets that have different numbers of classes.
We achieve FID scores of $5.85$ and $12.28$ on CIFAR10 and CIFAR100 respectively, which are comparable to state-of-the-art results. In addition to achieving good performance, our method has given us some inspiration. For example, we find the phenomenon that the coordination between the discriminator and generator is very important (\ie, to derive distinct class-aware generators, the discriminator must also be class-aware).
More interestingly, by analyzing the best model found by NAS, we find that the class-modulated convolution is more likely to appear in the early stage (close to the input noise) of the generator. We think this phenomenon is related to the semantic hierarchy of GANs~\citep{bau2018GANa, yang2020Semantic}. We apply this finding as an empirical rule to BigGAN~\citep{brock2018Large}, and also observe performance gain. This implies that our algorithm delivers useful and generalized insights to the design of cGAN models. We will release code and pre-trained models to facilitate future research.

\section{Related Work}

Generative Adversarial Network (GAN)~\citep{goodfellow2014Generative} have demonstrated impressive generation capabilities~\citep{karras2017Progressive,brock2018Large,karras2019StyleBased}. Nevertheless, it has notorious issues like vanishing gradient, training instability, and mode collapse. There are a number of improvements for the original GAN, \eg, changing the objective function~\citep{arjovsky2017Wasserstein,gulrajani2017Improved,mao2016Least,jolicoeur-martineau2019relativistic,qi2017LossSensitive}, improving network architecture~\citep{radford2015Unsupervised,brock2018Large,karras2019StyleBased,denton2015Deep,zhang2018SelfAttention,karnewar2019MSGGANa}, using multiple generators or discriminators~\citep{tolstikhin2017AdaGAN,hoang2018MGAN,arora2017Generalization,durugkar2017Generative,ghosh2018MultiAgent,nguyen2017dual}. Recently, the surge in neural architecture search (NAS) has triggered a wave of interest in automatically designing the network architecture of GAN~\citep{wang2019AGAN,gong2019AutoGAN,tian2020AlphaGAN,gao2019AdversarialNAS,tian2020OffPolicy,li2020GAN,fu2020AutoGANDistiller,kobayashi2020Multiobjective}.

Conditional GAN (cGAN)~\citep{mirza2014Conditional} is another type of GAN that incorporates class information into the original GAN, so that achieving promising results for the class-sensitive image generation task. Most of the early methods just incorporated the class information by concatenation~\citep{mirza2014Conditional,reed2016Generativea}. AC-GAN~\citep{odena2017Conditional} incorporated the label information into the objective function of the discriminator by an auxiliary classifier. \citet{miyato2018cGANs} proposed the class-projection (\textit{cproj}) discriminator, which injected class information into the discriminator in a projection-based way. Furthermore, conditional batch normalization (\textit{CBN})~\citep{devries2017Modulating} is a very effective method to modulate convolutional feature maps by conditional information. Subsequently, \textit{cproj} and \textit{CBN} are widely used together, forming some powerful cGANs for class image generation~\citep{zhang2018SelfAttention,brock2018Large}.

\section{Our Approach}


We use NAS to design class-aware generators for cGAN. However, implementing this cGAN model is not a trivial task. First, it is not easy to define the search space and the generator of each class may suffer insufficient training data. To tackle these issues, we detail the search space and a weight sharing strategy in Section~\ref{sec:search_space}. Second, we must design an efficient search method. In Section~\ref{sec:search_method}, we introduce the Multi-Net NAS (MN-NAS) and the \textit{mixed-architecture optimization}, these methods making the procedure of search and re-training of multiple networks simple.




\subsection{Search Space: Sharing Data by Class-Modulated Convolution}
\label{sec:search_space}

\begin{figure}[t]
   \centering
   \includegraphics[width=\linewidth]{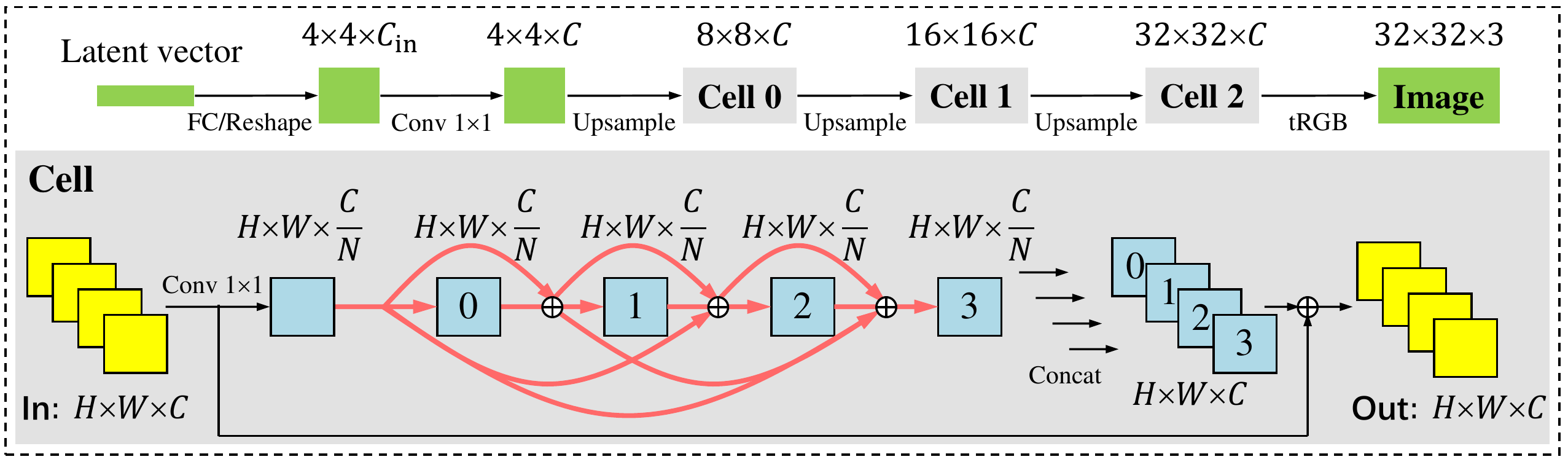}
   \caption{A tentative architecture in the search space. $N$ stands for the number of nodes in a cell, set to $4$ as an example. The operators shown in red arrows can be searched. The data shape is unchanged within each cell, so that the number of cells can be arbitrary. This figure is best viewed in color.}
   \label{fig:architecture}
\end{figure}

The design of the search space follows the popular cell-based style. The input latent vector is passed through $L$ up-sampling layers, each of which is followed by a cell that does not change the shape (spatial resolution and channel number) of the data. Each contains an input node, an output node, and $N$ intermediate nodes and there exists an edge between each pair of nodes, propagating neural responses from the lower-indexed node to the higher-indexed node. Each node summarizes inputs from its precedents, \ie, ${\mathbf{x}_j}={\sum_{i<j}o_{i,j}\left(\mathbf{x}_i\right)}$ where $o_{i,j}\left(\cdot\right)$ is the operator on edge $\left(i,j\right)$, chosen from the set of candidate operators, $\mathcal{O}$. To guarantee that the shape of data is unchanged, at the beginning of each cell, the data is pre-processed using a $1\times1$ convolutional layer that shrinks the number of channels by a factor of $N$. Hence, the output of intermediate nodes, after being concatenated, recover the original data shape. An architecture with tentative operators is shown in Figure~\ref{fig:architecture}.

Since the operator used in each edge can be searched, the number of different architectures is $\left|\mathcal{O}\right|^{L\times{N+1\choose2}}$. Note that we allow each class to have a distinct architecture, therefore, if there are $M$ classes in the dataset, the total number of possible combinations is $\left|\mathcal{O}\right|^{L\times{N+1\choose2}\times M}$. This is quite a large number. Even with the simplest setting used in this paper (\ie, ${\left|\mathcal{O}\right|}={2}$, ${L}={3}$, ${N}={2}$), this number is ${2^{90}}\approx{1.2\times10^{27}}$ for a $10$-class dataset (\eg, CIFAR10) or ${2^{900}}\approx{8.5\times10^{270}}$ for a $100$-class dataset (\eg, CIFAR100), much larger than some popular cell-based search spaces (\eg, the DARTS space~\citep{liu2019DARTS,chen2019Progressive,xu2020pc} with $1.1\times10^{18}$ architectures).

\begin{figure}[t]
   \centering
   \begin{minipage}{0.30\linewidth}
      \centering
      \includegraphics[width=\linewidth]{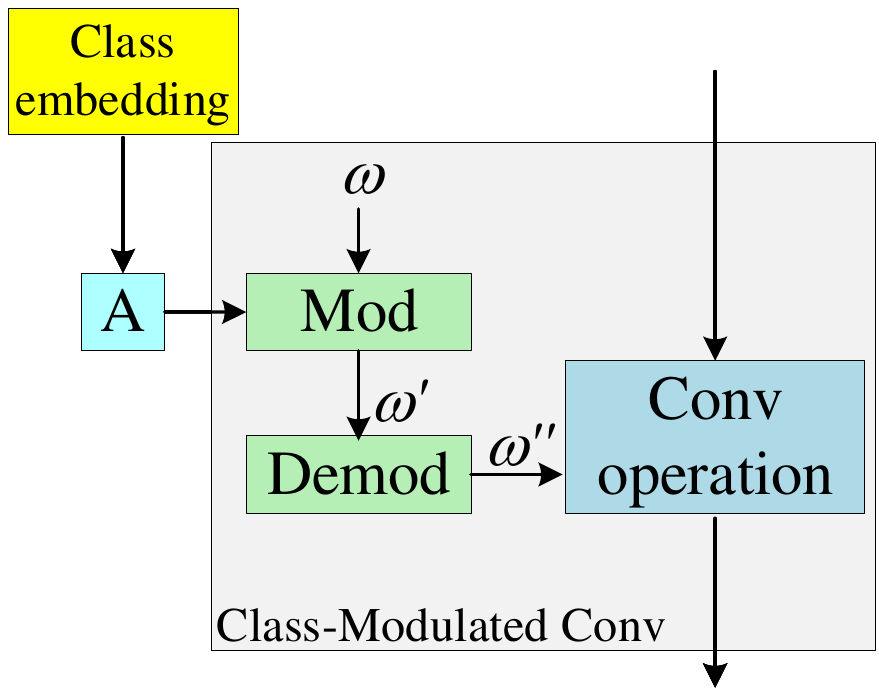}
      \caption{A class-modulated convolution (\textit{CMconv}), where the class-conditional vector (class embedding) is used to modulate the convolutional weights (shared with the regular convolution).}
      \label{fig:cmc}
   \end{minipage}%
   \hspace{0.8cm}
   \centering
   \begin{minipage}{0.50\linewidth}
      \centering
      \includegraphics[width=\linewidth]{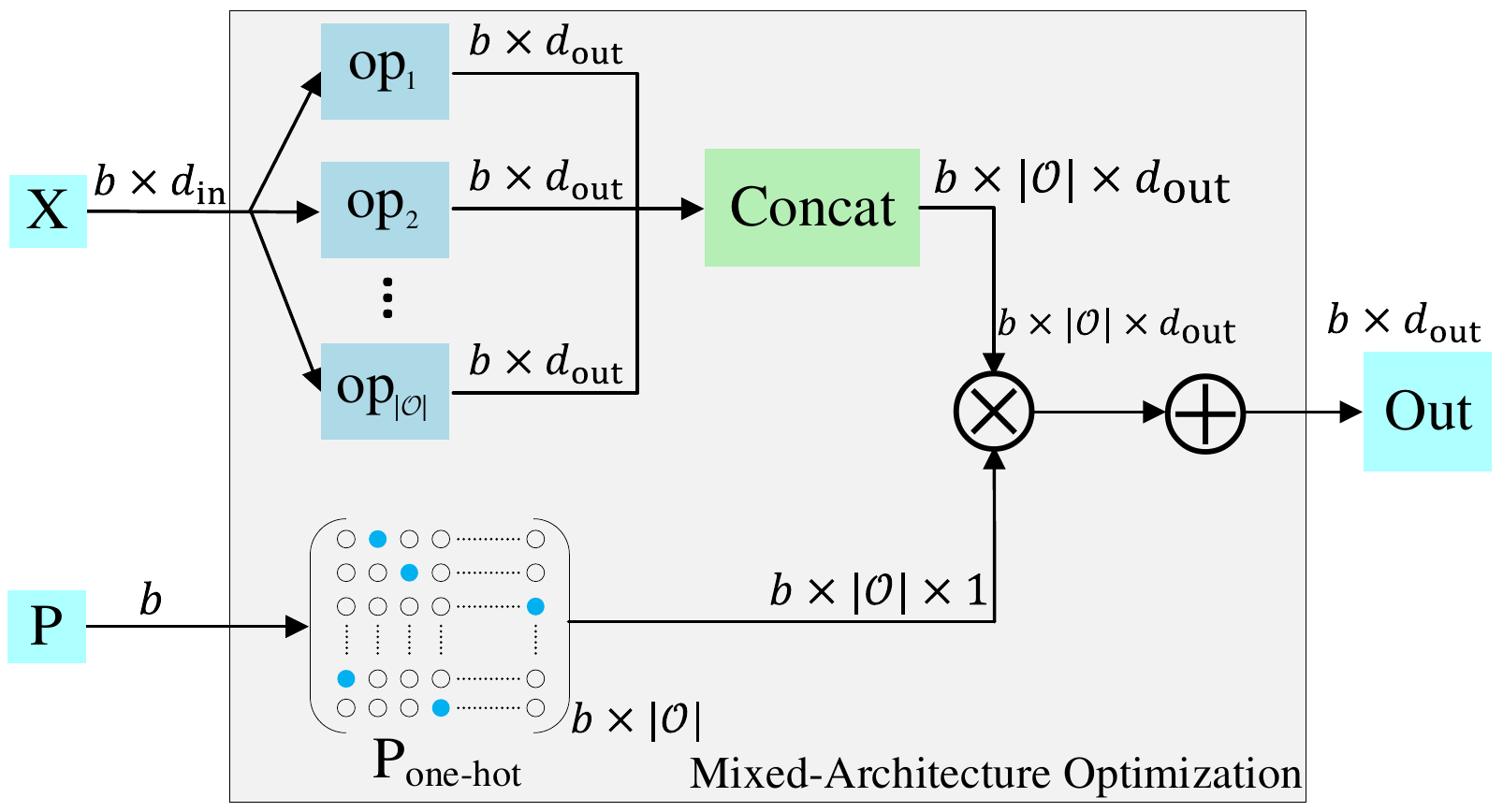}
      \caption{The mixed-architecture optimization with parallelization in a mini-batch. $\bigotimes$ stands for tensor-broadcasting multiplication, and $\bigoplus$ represents sum along the second dimension of the input}.
      \label{fig:mao}
   \end{minipage}
\end{figure}

\textbf{Class-Modulated Convolution.}\quad
The first challenge we encounter is that the training data need to be distributed among all classes. That being said, if the architectures of all classes are `truly' independent (\eg, all network weights are individually optimized), the amount of training data for each architecture, compared to the scenario that all classes share the same architecture, is reduced by a factor of $M$. This can result in severe over-fitting and unsatisfying performance of image generation.
To alleviate this risk, we share training data among different architectures by reusing model weights. In practice, most network weights are contributed by the convolutional layers, so we maintain one set of convolutional kernels and use a light-weighted module to introduce class-conditional information. Inspired by \textit{CBN}~\citep{devries2017Modulating} and the `demodulation' operation~\citep{karras2019Analyzing}, we propose the Class-Modulated convolution (\textit{CMconv}) operator to incorporate class-conditional information. As shown in Figure~\ref{fig:cmc}, a \textit{CMconv} layer consists of three parts, \textit{modulation}, \textit{demodulation}, and \textit{convolution}. The \textit{CMconv} shares convolutional weights with the corresponding regular convolution (\textit{Rconv}).

Mathematically, let $\mathbf{x}$ denote the input features maps with a class label of $y$, and $\boldsymbol{\omega}$ represent the weights of convolution. The goal of \textit{modulation} is to introduce a scale factor to each of the input channels, \ie, ${\boldsymbol{\omega}'}={\boldsymbol{\omega}\odot\mathbf{s}_\mathrm{in}}$ where both $\boldsymbol{\omega}$ and $\boldsymbol{\omega}'$ are in a shape of $c_\mathrm{in}\times c_\mathrm{out}\times U$. Here, $c_\mathrm{in}$ and $c_\mathrm{out}$ are the number of input and output channels, respectively, and $U$ is the kernel size (\eg, $3\times3$); $\mathbf{s}_\mathrm{in}$ is a $c_\mathrm{in}$-dimensional vector and $\boldsymbol{\omega}\odot\mathbf{s}_\mathrm{in}$ multiplies each set of weights ($c_\mathrm{out}\times U$ numbers) by the corresponding entry in $\mathbf{s}_\mathrm{in}$. We follow the conventional formulation of cGAN to define $\mathbf{s}_\mathrm{in}$ as an affine-transformed class vector, \ie, ${\mathbf{s}_\mathrm{in}}={\mathrm{Aff}\left(\mathbf{e}_y\right)}$, where $\mathrm{Aff}\left(\cdot\right)$ is simply implemented as a trainable fully-connected layer and $\mathbf{e}_y$ is a learnable embedding vector of the class label, $y$. The goal of \textit{demodulation} is to normalize the weights and keep the variance of the input and output features same. We follow~\citet{karras2019Analyzing} to use ${\boldsymbol{\omega}''}={\boldsymbol{\omega}'\odot\mathbf{s}_\mathrm{out}^{-1}}$ where $\mathbf{s}_\mathrm{out}$ is a $c_\mathrm{out}$-dimensional vector and $\mathbf{s}_\mathrm{out}^{-1}$ indicates element-wise reciprocal; $\mathbf{s}_\mathrm{out}$ is similar to the $\ell_2$-norm of $\boldsymbol{\omega}'$, computed as ${\mathbf{s}_\mathrm{out}}={\sqrt{\sum_{c_\mathrm{in},u}({\omega}'_{c_\mathrm{in},\cdot,u})^2+\epsilon}}$, where $\epsilon$ is a small constant to avoid numerical instability.

In summary, \textit{Rconv} and \textit{CMconv} start with the same weight, $\boldsymbol{\omega}$, and \textit{CMconv} modulates $\boldsymbol{\omega}$ into $\boldsymbol{\omega}''$. Then, regular convolution is performed, \ie, $\mathrm{conv}\left(\mathbf{x};\boldsymbol{\omega}\right)$ or $\mathrm{conv}\left(\mathbf{x};\boldsymbol{\omega}''\right)$. Since \textit{modulation} and \textit{demodulation} introduce relatively fewer parameters compared to convolution, so using weight-sharing \textit{Rconv} and \textit{CMconv} operators in each edge is a safe option that enables the limited amount of training data to be shared among a large number of generators.

\subsection{Search Method: Multi-Net NAS}
\label{sec:search_method}

The search process is formulated by a one-step Markov Decision Process (MDP). We denote $a$ as the action that samples architectures for all classes. Let ${\pi(a;\boldsymbol{\theta})}\in{\left(0,1\right)}$ be the sampling policy and $\boldsymbol{\theta}$ the learnable parameters, the performance of $\pi(a;\boldsymbol{\theta})$ is measured by:
\begin{equation}
   {J\left(\boldsymbol{\theta}\right)}={\mathbb{E}_{a\sim\pi\left(a;\boldsymbol{\theta}\right)}\left[R\left(a\right)\right]},
\end{equation}
where $R\left(a\right)$ is the reward function. Throughout this paper, we use the Inception Score as the reward. According to REINFORCE~\citep{williams1992Simple}, the gradient of $J\left(\boldsymbol{\theta}\right)$ with respect to $\boldsymbol{\theta}$ can be computed as:
\begin{equation}
   {\nabla_{\boldsymbol{\theta}}J\left(\boldsymbol{\theta}\right)}={\mathbb{E}_{a\sim\pi\left(a;\boldsymbol{\theta}\right)}\left[\left(R\left(a\right)-r\right)\cdot\nabla_{\boldsymbol{\theta}}\log\left(\pi\left(a;\boldsymbol{\theta}\right)\right)\right]}\approx{\frac{1}{m} \sum_{k=1}^m\left(R\left(a_k\right)-r\right)\cdot\nabla_{\boldsymbol{\theta}}\log\left(\pi\left(a_k;\boldsymbol{\theta}\right)\right)},
\end{equation}
where $m$ is the number of sampled architectures and $r$ is a baseline reward, set to be the moving average and used to reduce the variance in the training process. We use gradient ascent to maximize $J\left(\boldsymbol{\theta}\right)$. Inspired by~\citet{cai2019ProxylessNAS,ying2019NASBench101}, we design a simple policy. We use $\boldsymbol{\theta}_{l,k}$ to denote a $\left|\mathcal{O}\right|$-dimensional parameter of class $k$ and layer $l$, so that the probability of sampling each operator, $\mathrm{Prob}\left(o|\boldsymbol{\theta}_{l,k}\right)$, is determined by the softmax output of $\boldsymbol{\theta}_{l,k}$. Hence, given class-aware architectures, $A_{\mathrm{ca}}$, the probability that it gets sampled is ${\mathrm{Prob}\left(A_{\mathrm{ca}}|\pi\left(a;\boldsymbol{\theta}\right)\right)}={\prod_{l,k}\mathrm{Prob}\left(o|\boldsymbol{\theta}_{l,k}\right)}$.

\subsubsection{Overall Pipeline: Search, Re-Training, and Calibration}

There are two sets of parameters to be learned, \ie, $\boldsymbol{\theta}$ for the sampling policy and $\boldsymbol{\omega}$ for the super-network. We use the hinge adversarial loss~\citep{lim2017Geometric}:
\begin{equation}
   \begin{array}{rl}
      {\mathcal{L}_D} & ={\mathbb{E}_{q\left(y\right)}\left[\mathbb{E}_{q\left(\mathbf{x}|y\right)}\left[\max\{0,1-D\left(\mathbf{x},y\right\}\right]\right]+\mathbb{E}_{q\left(y\right)}\left[\mathbb{E}_{p\left(\mathbf{z}\right)}\left[\max\{0,1+D\left(G\left(\mathbf{z},y\right),y\right)\right]\right]}, \\
      {\mathcal{L}_G} & ={-\mathbb{E}_{q\left(y\right)}\left[\mathbb{E}_{p\left(\mathbf{z}\right)}\left[D\left(G\left(\mathbf{z},y\right),y\right)\right]\right]},
   \end{array}
\end{equation}
where $\mathbf{x}$ with class label $y$ is sampled from the real dataset, $D\left(\cdot\right)$ and $G\left(\cdot\right)$ denote the discriminator and generator, respectively, and $p\left(\mathbf{z}\right)$ is the standard Gaussian distribution. We use the discriminator in AutoGAN~\citep{gong2019AutoGAN} and add class projection (\textit{cproj}~\citep{miyato2018cGANs}) to it. We emphasize that using a class-aware discriminator is critical to our algorithm (please refer to Section~\ref{sec:cproj}).

During the \textbf{search} phase, we adopt the weight-sharing NAS approach~\citep{pham2018Efficient} to optimize the generator, \ie, the super-network parameterized by $\boldsymbol{\omega}$. We perform fair sampling strategy~\citep{chu2019FairNAS} to offer equal opportunity for training each generator architecture. After every ${T_\mathrm{critic}}={5}$ iterations, we update the generator weights; after every ${T_\mathrm{policy}}={50}$ iterations, we update the policy parameters, $\boldsymbol{\theta}$. The pseudo code of the optimization process is provided in Appendix \ref{apx-sec:search_algorithm}.

After the search, we obtain the generator architecture for each class by choosing the operator with the largest score on each edge, \ie, ${o_{l,k}}={\arg\max_o\mathrm{Prob}\left(o|\boldsymbol{\theta}_{l,k}\right)}$. Then, we \textbf{re-train} the generator from scratch following the same procedure. The last step is named \textbf{calibration}, in which we fine-tune the each architecture on the corresponding class for a small number of iterations (thus the overhead is small). As we will show in experiments, the calibration step largely boosts the performance, because the re-training stage has pursued for the optimality over all classes, which does not necessarily align with the optimality on each individual class.

\subsubsection{Sharing Computation by Mixed-Architecture Optimization}

We notice a technical issue that greatly downgrades the efficiency of both search and re-training. Given a mini-batch, $\mathcal{B}$, from the training set, as $\mathcal{B}$ may contain features from multiple classes, they need to be propagated through different architectures. To avoid heavy computational burden\footnote{If we deal with each class in a batch individually, the corresponding architectures need to be loaded to the GPU one by one and the batch size becomes small, which results in a reduction in computational efficiency. Moreover, inefficiency deteriorates with the number of classes increases.}, we propose \textit{mixed-architecture optimization} to allow different architectures to be optimized within one forward-then-backward pass in a batch.

The flowchart of mixed-architecture optimization is shown in Figure~\ref{fig:mao}. Let $\mathbf{X}$ denote a batch of input features of size $b\times d_\mathrm{in}$, where $b$ is the size of the batch, $\mathcal{B}$, and $d_\mathrm{in}$ is the feature dimensionality. To improve the efficiency of parallelization, $\mathbf{X}$ as a whole is propagated through every operator, and each training sample chooses the output corresponding to the selected operator. In practice, this is implemented by concatenating all the outputs of $o\left(\mathbf{X}\right)$, ${o}\in{\mathcal{O}}$, and multiply it by $\mathbf{P}$, a $b\times\left|\mathcal{O}\right|$ indicator matrix. Each row of which is a one-hot, $\left|\mathcal{O}\right|$-dimensional vector indicating the operator selected by the corresponding sample.

Essentially, mixed-architecture optimization performs redundant computation to achieve more efficient parallelization. Each input feature is fed into all $\left|\mathcal{O}\right|$ operators, though only one of the outputs will be used. Nevertheless, $b\times\left|\mathcal{O}\right|$ does not increase with the number of classes and thus our method generalizes well to complex datasets, \eg, CIFAR100.

\section{Experiments}


\textbf{Dataset and Evaluation.}\quad
We use CIFAR10 and CIFAR100~\citep{krizhevsky2009learning} as the testbeds. Both datasets have $50\rm{,}000$ training and $10\rm{,}000$ testing images, uniformly distributed over $10$ or $100$ classes. We use the Inception Score (IS)~\citep{salimans2016Improved} and the Fréchet Inception Distance (FID)~\citep{heusel2017GANs} to measure the performance of GAN on $50\mathrm{K}$ randomly generated images. The FID statistic files are pre-calculated using all training images. We also compute the FID score within each class of CIFAR10. Specifically, for each class, we first use $5\mathrm{K}$ real training images to pre-calculate the statistic file, and then randomly generate another $5\mathrm{K}$ images for computing the intra FID score. For more experimental details, please refer to Appendix~\ref{apx:details}.

\begin{table}[!t]
   \caption{FID scores of class-agnostic and class-aware GANs on CIFAR10.}
   \label{tab:cifar10}
   \resizebox{\textwidth}{!}{%
      \begin{tabular}{lccccccccccc}
         \toprule
         \multirow{2}{*}{Method}               & \multicolumn{10}{c}{Intra FIDs $\downarrow$} & \multicolumn{1}{l}{\multirow{2}{*}{FID $\downarrow$}}                                                                                                                                                                           \\ \cmidrule{2-11}
                                               & \textit{airp.}                               & \textit{auto.}                                        & \textit{bird}    & \textit{cat}     & \textit{deer}    & \textit{dog}     & \textit{frog}    & \textit{horse}   & \textit{ship}    & \textit{truck}   &                 \\
         \midrule
         \textsf{NAS-cGAN}                     & $29.10$                                      & $13.62$                                               & $26.64$          & $22.21$          & $14.97$          & $26.02$          & $17.32$          & $15.18$          & $14.99$          & $16.58$          & $7.05$          \\
         \textsf{NAS-cGAN} (calibrated)        & $26.83$                                      & $13.05$                                               & $25.99$          & $21.24$          & $14.56$          & $23.39$          & $17.32$          & $15.17$          & $14.46$          & $14.99$          & $6.63$          \\
         \midrule
         \textsf{NAS-caGAN}                    & $29.53$                                      & $12.32$                                               & $24.83$          & $21.30$          & $16.11$          & $26.64$          & $16.56$          & $16.79$          & $16.43$          & $16.39$          & $6.83$          \\
         \textsf{NAS-caGAN} (calibrated)       & $25.36$                                      & $\textbf{11.91}$                                      & $\textbf{22.66}$ & $\textbf{19.63}$ & $\textbf{13.74}$ & $23.29$          & $\textbf{15.81}$ & $15.60$          & $13.82$          & $\textbf{14.78}$ & $\textbf{5.85}$ \\
         \textsf{NAS-caGAN-light}              & $35.21$                                      & $12.64$                                               & $29.04$          & $24.96$          & $21.20$          & $26.26$          & $18.62$          & $16.40$          & $14.87$          & $18.67$          & $7.79$          \\
         \textsf{NAS-caGAN-light} (calibrated) & $\textbf{23.90}$                             & $12.69$                                               & $24.13$          & $23.36$          & $18.39$          & $\textbf{22.12}$ & $16.93$          & $\textbf{14.71}$ & $\textbf{13.17}$ & $15.01$          & $6.31$          \\
         \bottomrule
      \end{tabular}%
   }
\end{table}

\subsection{Quantitative Results}

\begin{figure}[!t]
   \centering
   \includegraphics[width=\linewidth]{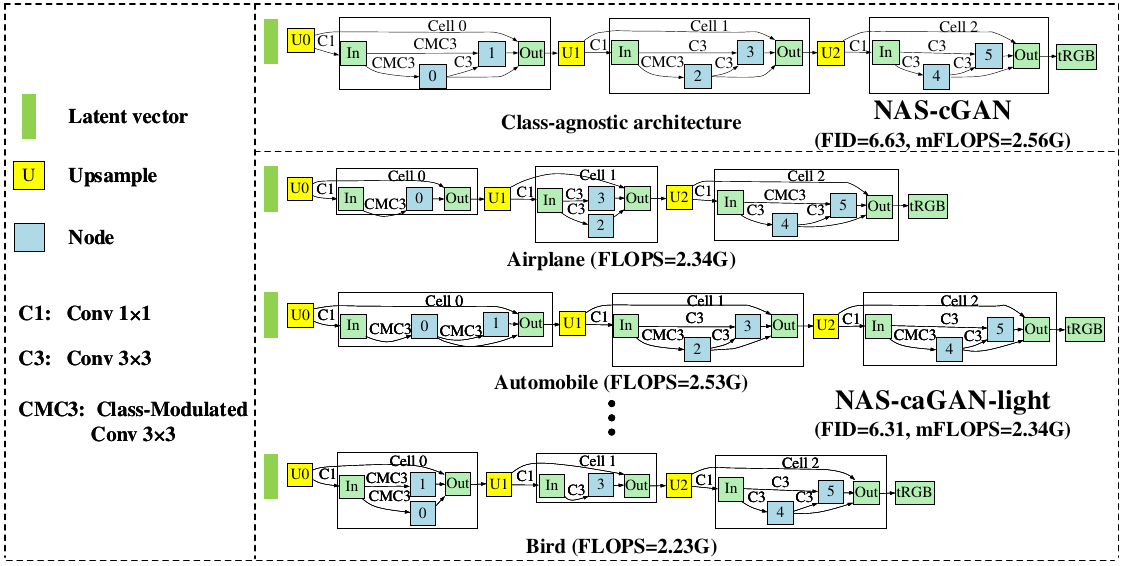}
   \caption{Part of the generator architectures found by \textsf{NAS-caGAN-light} on the CIFAR10 dataset. Compared to the class-agnostic architecture (shown at the top), the class-aware architectures enjoy a lower FID score as well as cheaper computation. This figure is best viewed in color.}
   \label{fig:visualization}
\end{figure}

Table~\ref{tab:cifar10} summarizes the image generation results on CIFAR10. We use \textsf{NAS-cGAN} to denote the method that each class uses the same searched generator architecture, and \textsf{NAS-caGAN} to denote the method with class-aware generators incorporated. The former option is achieved by a modified version of the proposed search algorithm that the sampling policy is shared among all classes. It can be seen that \textsf{NAS-caGAN} produces lower FID scores, indicating its better performance compared to \textsf{NAS-cGAN}. After calibration, \textsf{NAS-caGAN} achieves even better results, reporting an FID of $5.85$. We also compare the calibrated versions of \textsf{NAS-cGAN} and \textsf{NAS-caGAN} and find that the latter is better, indicating that both class-aware architectures and calibration contribute to the generating better images.

We notice that the current space chooses each operator between \textit{RConv} and \textit{CMConv}, both of which are parameterized and expensive. To find computationally efficient architectures, we add the \textit{zero} operator into the search space, and without any further modification, derive a light-weighted version of class-aware generators, denoted as \textsf{NAS-caGAN-light}. With or without calibration, \textsf{NAS-caGAN-light} achieves comparable FID values, while enjoying a reduced average computational overhead ($2.34\mathrm{G}$ FLOPs vs. $2.56\mathrm{G}$ FLOPs) of \textsf{NAS-caGAN}. This is another merit of using class-aware architectures. Figure~\ref{fig:visualization} shows part of the generator architectures found by \textsf{NAS-caGAN-light}. More architecture details are shown in Appendix~\ref{apx-sec:class_aware_arcs_cifar10}. Please refer to Appendix~\ref{apx:sec:comparison_cifar10} for comparison results with other cGAN models, and to Appendix~\ref{apx:sec:unconditional} for the unconditional image generation experiment.



Next, we challenge our method by evaluating it on CIFAR100 which has much more classes. Thanks to the proposed mixed-architecture optimization, we can perform architecture search on CIFAR100 without additional engineering efforts compared to that on CIFAR10. Differently, we do not perform calibration or evaluate the intra-class FID scores, since there are only $500$ training images for each class, which is insufficient to approximate the true image distribution.

\begin{wraptable}{r}{8cm}
   \vspace{-0.48cm}
   \centering
   \caption{FID and IS scores on CIFAR100. $^\dagger$ indicates quoted from the paper.}
   \begin{tabular}{lcc}
      \toprule
      Method                                 & FID $\downarrow$ & IS $\uparrow$    \\
      \midrule
      SN-GAN~\citep{miyato2018Spectral}      & $18.87$          & $8.19$           \\
      \textit{cproj}~\citep{miyato2018cGANs} & $23.20^\dagger$  & $9.04^\dagger$   \\
      Multi-hinge~\citep{kavalerov2019cGANs} & $14.62$          & $\textbf{13.35}$ \\
      FQ-GAN~\citep{zhao2020Feature}         & $\textbf{8.23}$  & $10.62$          \\
      \midrule
      \textsf{NAS-cGAN} (ours)               & $13.94$          & $8.83$           \\
      \textsf{NAS-caGAN} (ours)              & $12.28$          & $9.71$           \\
      \bottomrule
   \end{tabular}
   \label{tab:cifar100}
\end{wraptable}

Still, we use \textsf{NAS-cGAN} and \textsf{NAS-caGAN} to denote the class-agnostic and class-aware versions of cGAN, respectively. Table~\ref{tab:cifar100} summarizes the results. Again, by allowing different classes to have individual generator architectures, \textsf{NAS-caGAN} achieves better performance in terms of both the FID and IS scores. The searched architectures are shown in Appendix~\ref{apx-sec:class_aware_arcs_cifar100}. Though we have not achieved the state-of-the-art results, the idea of designing class-aware generators indeed brings benefits. We believe that our findings can be incorporated into other methods (\eg, Multi-hinge~\citep{kavalerov2019cGANs}, FQ-GAN~\citep{zhao2020Feature}) for better performance.


\subsection{Diagnostic Studies}
\label{sec:analysis}


\subsubsection{Coordination between Discriminator and Class-Aware Generators}
\label{sec:cproj}

\begin{figure}[!t]
   \centering
   \begin{subfigure}{0.30\textwidth}
      \centering
      \includegraphics[width=\linewidth]{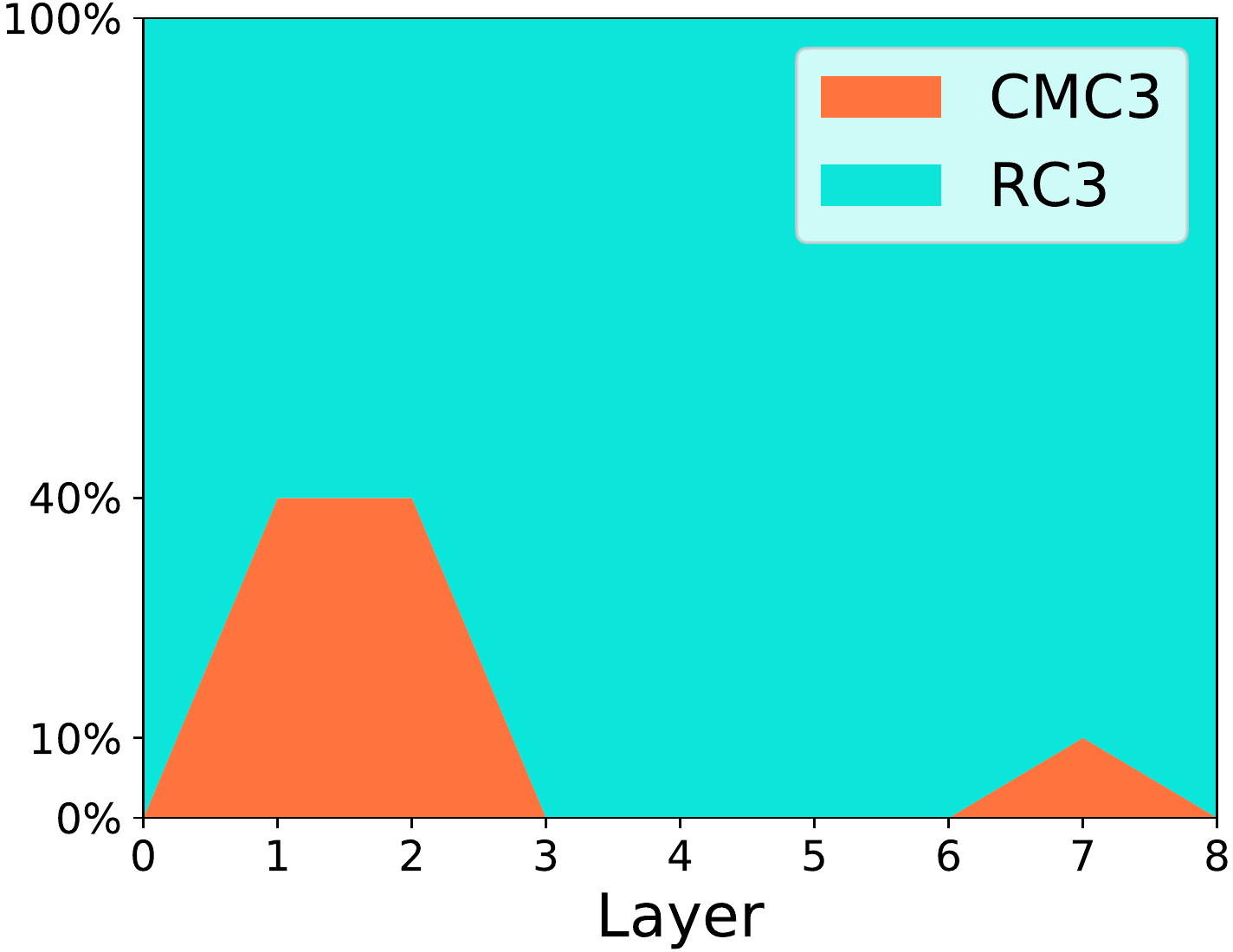}
      \caption{}
      \label{fig:normal_disc}
   \end{subfigure}
   \begin{subfigure}{0.30\textwidth}
      \centering
      \includegraphics[width=\linewidth]{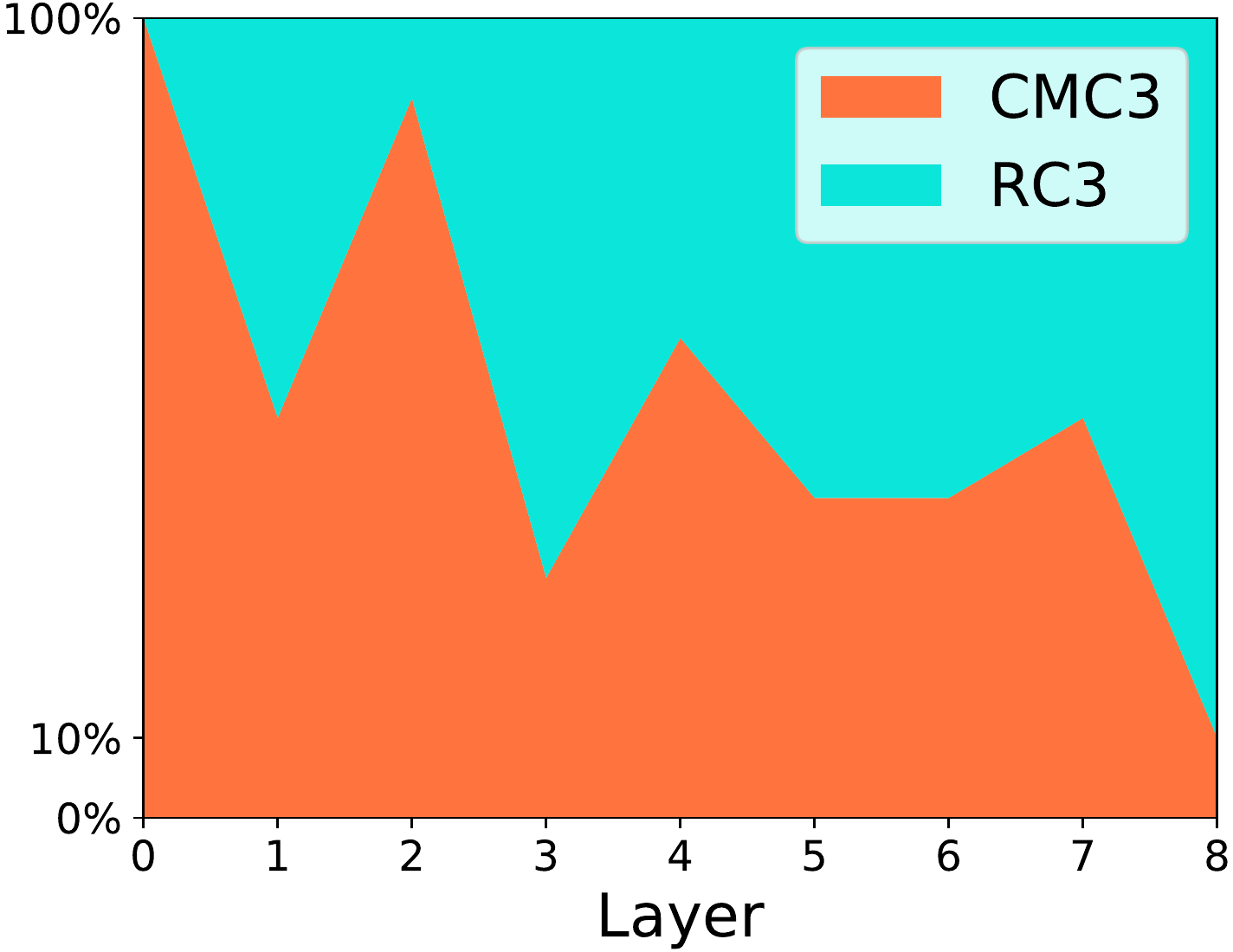}
      \caption{}
      \label{fig:cproj_cifar10}
   \end{subfigure}
   \begin{subfigure}{.30\textwidth}
      \centering
      \includegraphics[width=\linewidth]{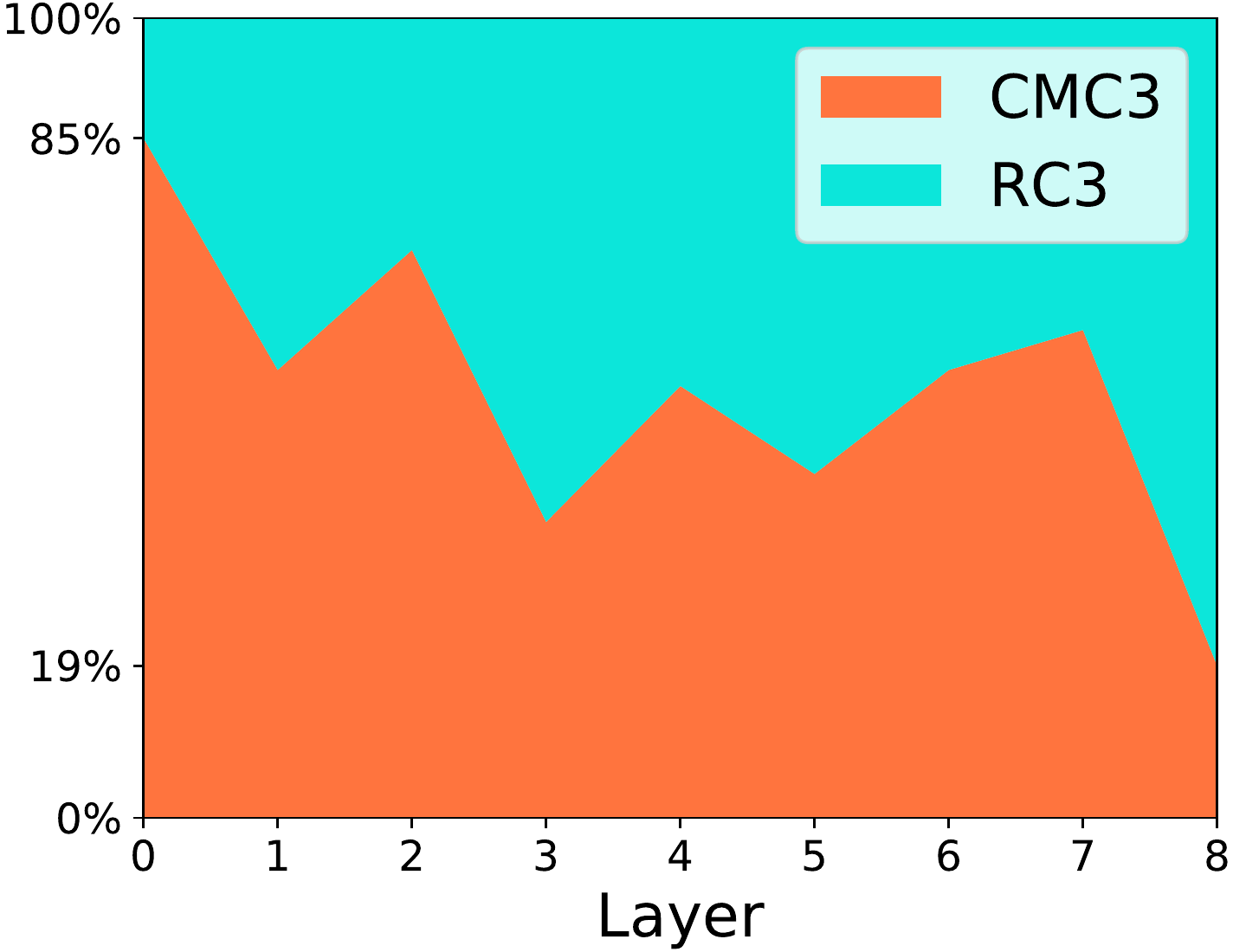}
      \caption{}
      \label{fig:cproj_cifar100}
   \end{subfigure}
   \vspace{-0.2cm}
   \caption{The proportion of \textit{RConv} and \textit{CMConv} in each layer of the searched class-aware architectures, where (a) is obtained using a normal discriminator on CIFAR10, (b) is obtained using a \textit{cproj} discriminator on CIFAR10, and (c) is obtained using a \textit{cproj} discriminator on CIFAR100. See Section~\ref{sec:analysis} for details.}
   \label{fig:fig}
\end{figure}

Based on the \textsf{NAS-caGAN} model, we investigate the difference between using a normal discriminator (\ie, no class information) and using a class-projection (\textit{cproj}) discriminator~\citep{miyato2018cGANs}. On CIFAR10, these models report FID scores of $15.90$ and $6.83$, respectively, \ie, the \textit{cproj} discriminator is significantly better. We owe the performance gain to the ability that \textit{cproj} induces more \textit{CMConv} operators. As shown in Figure~\ref{fig:normal_disc}, the normal discriminator leads to a sparse use of \textit{CMConv} operators, while, as in Figure~\ref{fig:cproj_cifar10}, \textit{CMConv} occupies almost half of the operators when \textit{cproj} is used. That is, the class-aware generators should be searched under the condition that the discriminator is also class-aware. This aligns with the results obtained in~\citep{miyato2018cGANs}, showing that one usually uses the \textit{cproj} discriminator together with the generator containing conditional batch normalization operations.

\subsubsection{Where Shall We Place Class-Aware Operators?}


Last but not least, we study how our algorithm assigns the class-aware operator (\ie, \textit{CMConv}) to different positions of the searched generator architectures. Continuing the previous study, we plot the portion of \textit{CMConv} on CIFAR100 in Figure~\ref{fig:cproj_cifar100}, which shows a very similar trend as the experiments on CIFAR10. In particular, the portion for the first operator (close to the input noise vector) to use \textit{CMConv} is $100\%$ on CIFAR10 and $85\%$ on CIFAR100, while the portion for the last operator (close to the output generated image) is only $10\%$ and $19\%$ on CIFAR10 and CIFAR100, respectively. This is to suggest that class information seems very important for capturing the distribution of high-level semantics (close to input), while all classes seem to share similar low-level patterns (close to output).

To verify that the finding can indeed enhance GAN models, we experiment on BigGAN~\citep{brock2018Large}, a manually designed GAN model. The generator of BigGAN uses three blocks on CIFAR10, each of which contains two class-conditional operations (\ie, \textit{CBN}) by default.
We try to put the \textit{CBN} in different blocks to study the relationship between accuracy and injection position of class information. As shown in Table~\ref{tab:cbn_cmc_position}, the model using \textit{CBN} only in the first block (index $0$, close to the input noise) works better than those using \textit{CBN} in other blocks and even the original BigGAN model (using \textit{CBN} in all blocks). And using \textit{CBN} only in the last block (close to output) produces inferior performance. Please refer to Appendix~\ref{apx:cbn_cmconv} for more details.

\begin{wraptable}{r}{5cm}
   \vspace{-0.45cm}
   \caption{Evaluating the BigGAN architecture on CIFAR10 with \textit{CBN} (or \textit{CMConv}) inserted into different blocks. The table is sorted by the `\textit{CMConv}' column. Please refer to the main texts for details.}
   \vspace{-0.3cm}
   \label{tab:cbn_cmc_position}
   \resizebox{\linewidth}{!}{%
      \begin{tabular}{ccccc}
         \toprule
         \multicolumn{3}{c}{Block index} & \multirow{2}{*}{\textit{CBN}} & \multirow{2}{*}{\textit{CMConv}}                                     \\ \cmidrule{1-3}
         0                               & 1                             & 2                                &                 &                 \\ \midrule
         \ding{52}                       &                               &                                  & $\textbf{7.11}$ & $\textbf{6.91}$ \\
         \ding{52}                       & \ding{52}                     &                                  & $7.39$          & $7.00$          \\
         \ding{52}                       & \ding{52}                     & \ding{52}                        & $7.55$          & $7.15$          \\
         \ding{52}                       &                               & \ding{52}                        & $7.28$          & $7.29$          \\
                                         & \ding{52}                     & \ding{52}                        & $8.02$          & $8.06$          \\
                                         & \ding{52}                     &                                  & $7.97$          & $8.15$          \\
                                         &                               & \ding{52}                        & $10.60$         & $9.96$          \\
         \bottomrule
      \end{tabular}%
   }
\end{wraptable}

Next, we study the relationship between accuracy and the position of \textit{CMConv} by replacing the \textit{CBN} in BigGAN with regular \textit{BN}, and replacing the regular convolution with \textit{CMConv}, with other hyperparameters consistent with BigGAN.  The results in Table~\ref{tab:cbn_cmc_position} show a similar trend to the experiments of \textit{CBN}. Besides, the best model using \textit{CMConv} in the first block is slightly better than the best model of \textit{CBN} ($6.91$ vs. $7.11$).

We explain this phenomenon by the finding~\citep{bau2018GANa,yang2020Semantic} that for the generator of GAN, the early and the middle layers determine the spatial layout and the category attributes, so the earlier the injection of category information, the more benefits; the final layers control the lighting and color scheme of the generated images, which are shared attributes for different classes, so using the regular convolutional layer is fine. Interestingly, our algorithm offers an alternative way to reveal this rule which was previously discovered by human experts. This indicates that our method is helpful for understanding and improving existing hand-designed generator network architectures.



\section{Conclusions}

In this paper, we reveal the possibility of designing class-aware generators for conditional GAN models. Though the motivation seems straightforward, non-trivial technical efforts are required to improve the performance as well as efficiency of the algorithm, in particular when the number of classes becomes large. We claim two key contributions. \textbf{First}, the search space containing regular and class-modulated convolutions eases the re-training process because all convolutional weights can be trained on the full dataset. \textbf{Second}, we design a mixed-architecture optimization mechanism so that the search and re-training processes can be performed efficiently. In a relatively small search space, the proposed Multi-Net NAS (MN-NAS) algorithm achieves FID scores of $5.85$ and $12.28$ on CIFAR10 and CIFAR100, respectively. Provided more computational resources, our algorithm has the potential of generating images of higher quality.

Our research leaves some open problems to the community. First, we have used the constraint that all operators are either regular or class-modulated convolution. This is to improve the efficiency of utilizing training data, but this also limits the diversity of the searched generator architectures. Second, it is interesting to jointly optimize the architectures of generator and discriminator, since we have found strong evidence of their cooperation. We leave these topics to future work.


\bibliography{iclr2021_conference}

\begin{thebibliography}{53}
\providecommand{\natexlab}[1]{#1}
\providecommand{\url}[1]{\texttt{#1}}
\expandafter\ifx\csname urlstyle\endcsname\relax
  \providecommand{\doi}[1]{doi: #1}\else
  \providecommand{\doi}{doi: \begingroup \urlstyle{rm}\Url}\fi

\bibitem[Arjovsky et~al.(2017)Arjovsky, Chintala, and
  Bottou]{arjovsky2017Wasserstein}
Martin Arjovsky, Soumith Chintala, and L{\'e}on Bottou.
\newblock Wasserstein {{Generative Adversarial Networks}}.
\newblock In \emph{{{ICML}}}, 2017.

\bibitem[Arora et~al.(2017)Arora, Ge, Liang, Ma, and
  Zhang]{arora2017Generalization}
Sanjeev Arora, Rong Ge, Yingyu Liang, Tengyu Ma, and Yi~Zhang.
\newblock Generalization and {{Equilibrium}} in {{Generative Adversarial Nets}}
  ({{GANs}}).
\newblock \emph{arXiv:1703.00573 [cs, stat]}, 2017.

\bibitem[Bau et~al.(2018)Bau, Zhu, Strobelt, Zhou, Tenenbaum, Freeman, and
  Torralba]{bau2018GANa}
David Bau, Jun-Yan Zhu, Hendrik Strobelt, Bolei Zhou, Joshua~B. Tenenbaum,
  William~T. Freeman, and Antonio Torralba.
\newblock {{GAN Dissection}}: {{Visualizing}} and {{Understanding Generative
  Adversarial Networks}}.
\newblock \emph{arXiv:1811.10597 [cs]}, 2018.

\bibitem[Brock et~al.(2018)Brock, Donahue, and Simonyan]{brock2018Large}
Andrew Brock, Jeff Donahue, and Karen Simonyan.
\newblock Large scale gan training for high fidelity natural image synthesis.
\newblock \emph{arXiv:1809.11096}, 2018.

\bibitem[Cai et~al.(2019)Cai, Zhu, and Han]{cai2019ProxylessNAS}
Han Cai, Ligeng Zhu, and Song Han.
\newblock {{ProxylessNAS}}: {{Direct Neural Architecture Search}} on {{Target
  Task}} and {{Hardware}}.
\newblock In \emph{{{ICLR}}}, 2019.

\bibitem[Chen et~al.(2019)Chen, Xie, Wu, and Tian]{chen2019Progressive}
Xin Chen, Lingxi Xie, Jun Wu, and Qi~Tian.
\newblock Progressive {{Differentiable Architecture Search}}: {{Bridging}} the
  {{Depth Gap}} between {{Search}} and {{Evaluation}}.
\newblock In \emph{{{ICCV}}}, 2019.

\bibitem[Chen et~al.(2020)Chen, Xie, Wu, Wei, Xu, and Tian]{chen2020Fitting}
Xin Chen, Lingxi Xie, Jun Wu, Longhui Wei, Yuhui Xu, and Qi~Tian.
\newblock Fitting the {{Search Space}} of {{Weight}}-sharing {{NAS}} with
  {{Graph Convolutional Networks}}.
\newblock \emph{arXiv:2004.08423 [cs, stat]}, 2020.

\bibitem[Chu et~al.(2019)Chu, Zhang, Xu, and Li]{chu2019FairNAS}
Xiangxiang Chu, Bo~Zhang, Ruijun Xu, and Jixiang Li.
\newblock {{FairNAS}}: {{Rethinking Evaluation Fairness}} of {{Weight Sharing
  Neural Architecture Search}}.
\newblock \emph{arXiv:1907.01845 [cs, stat]}, 2019.

\bibitem[{de Vries} et~al.(2017){de Vries}, Strub, Mary, Larochelle, Pietquin,
  and Courville]{devries2017Modulating}
Harm {de Vries}, Florian Strub, J{\'e}r{\'e}mie Mary, Hugo Larochelle, Olivier
  Pietquin, and Aaron Courville.
\newblock Modulating early visual processing by language.
\newblock In \emph{{{NeurIPS}}}, 2017.

\bibitem[Denton et~al.(2015)Denton, Chintala, Szlam, and
  Fergus]{denton2015Deep}
Emily Denton, Soumith Chintala, Arthur Szlam, and Rob Fergus.
\newblock Deep {{Generative Image Models}} using a {{Laplacian Pyramid}} of
  {{Adversarial Networks}}.
\newblock \emph{arXiv:1506.05751 [cs]}, 2015.

\bibitem[Durugkar et~al.(2017)Durugkar, Gemp, and
  Mahadevan]{durugkar2017Generative}
Ishan Durugkar, Ian Gemp, and Sridhar Mahadevan.
\newblock Generative {{Multi}}-{{Adversarial Networks}}.
\newblock In \emph{{{ICLR}}}, 2017.

\bibitem[Fu et~al.(2020)Fu, Chen, Wang, Li, Lin, and
  Wang]{fu2020AutoGANDistiller}
Yonggan Fu, Wuyang Chen, Haotao Wang, Haoran Li, Yingyan Lin, and Zhangyang
  Wang.
\newblock {{AutoGAN}}-{{Distiller}}: {{Searching}} to {{Compress Generative
  Adversarial Networks}}.
\newblock In \emph{{{ICML}}}, 2020.

\bibitem[Gao et~al.(2019)Gao, Chen, Liu, Tan, and Yan]{gao2019AdversarialNAS}
Chen Gao, Yunpeng Chen, Si~Liu, Zhenxiong Tan, and Shuicheng Yan.
\newblock {{AdversarialNAS}}: {{Adversarial Neural Architecture Search}} for
  {{GANs}}.
\newblock \emph{arXiv:1912.02037 [cs, eess]}, 2019.

\bibitem[Ghosh et~al.(2018)Ghosh, Kulharia, Namboodiri, Torr, and
  Dokania]{ghosh2018MultiAgent}
Arnab Ghosh, Viveka Kulharia, Vinay Namboodiri, Philip H.~S. Torr, and
  Puneet~K. Dokania.
\newblock Multi-{{Agent Diverse Generative Adversarial Networks}}.
\newblock In \emph{{{CVPR}}}, 2018.

\bibitem[Gong et~al.(2019)Gong, Chang, Jiang, and Wang]{gong2019AutoGAN}
Xinyu Gong, Shiyu Chang, Yifan Jiang, and Zhangyang Wang.
\newblock {{AutoGAN}}: {{Neural Architecture Search}} for {{Generative
  Adversarial Networks}}.
\newblock In \emph{{{ICCV}}}, 2019.

\bibitem[Goodfellow et~al.(2014)Goodfellow, {Pouget-Abadie}, Mirza, Xu,
  {Warde-Farley}, Ozair, Courville, and Bengio]{goodfellow2014Generative}
Ian Goodfellow, Jean {Pouget-Abadie}, Mehdi Mirza, Bing Xu, David
  {Warde-Farley}, Sherjil Ozair, Aaron Courville, and Yoshua Bengio.
\newblock Generative {{Adversarial Nets}}.
\newblock In \emph{{{NeurIPS}}}, 2014.

\bibitem[Gulrajani et~al.(2017)Gulrajani, Ahmed, Arjovsky, Dumoulin, and
  Courville]{gulrajani2017Improved}
Ishaan Gulrajani, Faruk Ahmed, Martin Arjovsky, Vincent Dumoulin, and Aaron~C
  Courville.
\newblock Improved training of wasserstein gans.
\newblock In \emph{{{NeurIPS}}}, 2017.

\bibitem[Heusel et~al.(2017)Heusel, Ramsauer, Unterthiner, Nessler, and
  Hochreiter]{heusel2017GANs}
Martin Heusel, Hubert Ramsauer, Thomas Unterthiner, Bernhard Nessler, and Sepp
  Hochreiter.
\newblock {{GANs Trained}} by a {{Two Time}}-{{Scale Update Rule Converge}} to
  a {{Local Nash Equilibrium}}.
\newblock In \emph{{{NeurIPS}}}, 2017.

\bibitem[Hoang et~al.(2018)Hoang, Nguyen, Le, and Phung]{hoang2018MGAN}
Quan Hoang, Tu~Dinh Nguyen, Trung Le, and Dinh Phung.
\newblock {{MGAN}}: {{Training Generative Adversarial Nets}} with {{Multiple
  Generators}}.
\newblock In \emph{{{ICLR}}}, 2018.

\bibitem[{Jolicoeur-Martineau}(2019)]{jolicoeur-martineau2019relativistic}
Alexia {Jolicoeur-Martineau}.
\newblock The relativistic discriminator: A key element missing from standard
  {{GAN}}.
\newblock In \emph{{{ICLR}}}, 2019.

\bibitem[Karnewar \& Wang(2019)Karnewar and Wang]{karnewar2019MSGGANa}
Animesh Karnewar and Oliver Wang.
\newblock {{MSG}}-{{GAN}}: {{Multi}}-{{Scale Gradient GAN}} for {{Stable Image
  Synthesis}}.
\newblock \emph{arXiv:1903.06048 [cs, stat]}, 2019.

\bibitem[Karras et~al.(2017)Karras, Aila, Laine, and
  Lehtinen]{karras2017Progressive}
Tero Karras, Timo Aila, Samuli Laine, and Jaakko Lehtinen.
\newblock Progressive {{Growing}} of {{GANs}} for {{Improved Quality}},
  {{Stability}}, and {{Variation}}.
\newblock \emph{arXiv:1710.10196 [cs, stat]}, 2017.

\bibitem[Karras et~al.(2019{\natexlab{a}})Karras, Laine, and
  Aila]{karras2019StyleBased}
Tero Karras, Samuli Laine, and Timo Aila.
\newblock A {{Style}}-{{Based Generator Architecture}} for {{Generative
  Adversarial Networks}}.
\newblock In \emph{{{CVPR}}}, 2019{\natexlab{a}}.

\bibitem[Karras et~al.(2019{\natexlab{b}})Karras, Laine, Aittala, Hellsten,
  Lehtinen, and Aila]{karras2019Analyzing}
Tero Karras, Samuli Laine, Miika Aittala, Janne Hellsten, Jaakko Lehtinen, and
  Timo Aila.
\newblock Analyzing and {{Improving}} the {{Image Quality}} of {{StyleGAN}}.
\newblock \emph{arXiv:1912.04958 [cs, eess, stat]}, 2019{\natexlab{b}}.

\bibitem[Karras et~al.(2020)Karras, Aittala, Hellsten, Laine, Lehtinen, and
  Aila]{karras2020Training}
Tero Karras, Miika Aittala, Janne Hellsten, Samuli Laine, Jaakko Lehtinen, and
  Timo Aila.
\newblock Training {{Generative Adversarial Networks}} with {{Limited Data}}.
\newblock \emph{arXiv:2006.06676 [cs, stat]}, 2020.

\bibitem[Kavalerov et~al.(2019)Kavalerov, Czaja, and
  Chellappa]{kavalerov2019cGANs}
Ilya Kavalerov, Wojciech Czaja, and Rama Chellappa.
\newblock {{cGANs}} with {{Multi}}-{{Hinge Loss}}.
\newblock \emph{arXiv:1912.04216 [cs, stat]}, 2019.

\bibitem[Kingma \& Ba(2014)Kingma and Ba]{kingma2014Adam}
Diederik~P. Kingma and Jimmy Ba.
\newblock Adam: {{A Method}} for {{Stochastic Optimization}}.
\newblock In \emph{{{ICLR}}}, 2014.

\bibitem[Kobayashi \& Nagao(2020)Kobayashi and
  Nagao]{kobayashi2020Multiobjective}
Masayuki Kobayashi and Tomoharu Nagao.
\newblock A {{Multi}}-objective architecture search for generative adversarial
  networks.
\newblock In \emph{Proceedings of the 2020 {{Genetic}} and {{Evolutionary
  Computation Conference Companion}}}, 2020.

\bibitem[Krizhevsky et~al.(2009)Krizhevsky, Hinton,
  et~al.]{krizhevsky2009learning}
Alex Krizhevsky, Geoffrey Hinton, et~al.
\newblock Learning multiple layers of features from tiny images.
\newblock Technical report, {Citeseer}, 2009.

\bibitem[Li et~al.(2020)Li, Lin, Ding, Liu, Zhu, and Han]{li2020GAN}
Muyang Li, Ji~Lin, Yaoyao Ding, Zhijian Liu, Jun-Yan Zhu, and Song Han.
\newblock {{GAN Compression}}: {{Efficient Architectures}} for {{Interactive
  Conditional GANs}}.
\newblock In \emph{{{CVPR}}}, 2020.

\bibitem[Lim \& Ye(2017)Lim and Ye]{lim2017Geometric}
Jae~Hyun Lim and Jong~Chul Ye.
\newblock Geometric {{GAN}}.
\newblock \emph{arXiv:1705.02894 [cond-mat, stat]}, 2017.

\bibitem[Liu et~al.(2019)Liu, Simonyan, and Yang]{liu2019DARTS}
Hanxiao Liu, Karen Simonyan, and Yiming Yang.
\newblock {{DARTS}}: {{Differentiable Architecture Search}}.
\newblock In \emph{{{ICLR}}}, 2019.

\bibitem[Mao et~al.(2016)Mao, Li, Xie, Lau, Wang, and Smolley]{mao2016Least}
Xudong Mao, Qing Li, Haoran Xie, Raymond Y.~K. Lau, Zhen Wang, and Stephen~Paul
  Smolley.
\newblock Least {{Squares Generative Adversarial Networks}}.
\newblock \emph{arXiv:1611.04076 [cs]}, 2016.

\bibitem[Mirza \& Osindero(2014)Mirza and Osindero]{mirza2014Conditional}
Mehdi Mirza and Simon Osindero.
\newblock Conditional {{Generative Adversarial Nets}}.
\newblock \emph{arXiv:1411.1784 [cs, stat]}, 2014.

\bibitem[Miyato \& Koyama(2018)Miyato and Koyama]{miyato2018cGANs}
Takeru Miyato and Masanori Koyama.
\newblock {{cGANs}} with {{Projection Discriminator}}.
\newblock In \emph{{{ICLR}}}, 2018.

\bibitem[Miyato et~al.(2018)Miyato, Kataoka, Koyama, and
  Yoshida]{miyato2018Spectral}
Takeru Miyato, Toshiki Kataoka, Masanori Koyama, and Yuichi Yoshida.
\newblock Spectral {{Normalization}} for {{Generative Adversarial Networks}}.
\newblock \emph{arXiv:1802.05957 [cs, stat]}, 2018.

\bibitem[Nguyen et~al.(2017)Nguyen, Le, Vu, and Phung]{nguyen2017dual}
Tu~Nguyen, Trung Le, Hung Vu, and Dinh Phung.
\newblock Dual discriminator generative adversarial nets.
\newblock In \emph{{{NeurIPS}}}, 2017.

\bibitem[Odena et~al.(2017)Odena, Olah, and Shlens]{odena2017Conditional}
Augustus Odena, Christopher Olah, and Jonathon Shlens.
\newblock Conditional {{Image Synthesis With Auxiliary Classifier GANs}}.
\newblock \emph{arXiv:1610.09585 [cs, stat]}, 2017.

\bibitem[Pham et~al.(2018)Pham, Guan, Zoph, Le, and Dean]{pham2018Efficient}
Hieu Pham, Melody~Y. Guan, Barret Zoph, Quoc~V. Le, and Jeff Dean.
\newblock Efficient {{Neural Architecture Search}} via {{Parameter Sharing}}.
\newblock In \emph{{{ICML}}}, 2018.

\bibitem[Qi(2017)]{qi2017LossSensitive}
Guo-Jun Qi.
\newblock Loss-{{Sensitive Generative Adversarial Networks}} on {{Lipschitz
  Densities}}.
\newblock \emph{arXiv:1701.06264 [cs]}, 2017.

\bibitem[Radford et~al.(2015)Radford, Metz, and
  Chintala]{radford2015Unsupervised}
Alec Radford, Luke Metz, and Soumith Chintala.
\newblock Unsupervised {{Representation Learning}} with {{Deep Convolutional
  Generative Adversarial Networks}}.
\newblock \emph{arXiv:1511.06434 [cs]}, 2015.

\bibitem[Reed et~al.(2016)Reed, Akata, Yan, Logeswaran, Schiele, and
  Lee]{reed2016Generativea}
Scott Reed, Zeynep Akata, Xinchen Yan, Lajanugen Logeswaran, Bernt Schiele, and
  Honglak Lee.
\newblock Generative {{Adversarial Text}} to {{Image Synthesis}}.
\newblock In \emph{{{ICML}}}, 2016.

\bibitem[Salimans et~al.(2016)Salimans, Goodfellow, Zaremba, Cheung, Radford,
  Chen, and Chen]{salimans2016Improved}
Tim Salimans, Ian Goodfellow, Wojciech Zaremba, Vicki Cheung, Alec Radford,
  Xi~Chen, and Xi~Chen.
\newblock Improved {{Techniques}} for {{Training GANs}}.
\newblock In \emph{{{NeurIPS}}}, 2016.

\bibitem[Tian et~al.(2020{\natexlab{a}})Tian, Wang, Huang, Li, Dai, Yang, Wang,
  and Fink]{tian2020OffPolicy}
Yuan Tian, Qin Wang, Zhiwu Huang, Wen Li, Dengxin Dai, Minghao Yang, Jun Wang,
  and Olga Fink.
\newblock Off-{{Policy Reinforcement Learning}} for {{Efficient}} and
  {{Effective GAN Architecture Search}}.
\newblock In \emph{{{ECCV}}}, 2020{\natexlab{a}}.

\bibitem[Tian et~al.(2020{\natexlab{b}})Tian, Shen, Shen, Su, Li, and
  Liu]{tian2020AlphaGAN}
Yuesong Tian, Li~Shen, Li~Shen, Guinan Su, Zhifeng Li, and Wei Liu.
\newblock {{AlphaGAN}}: {{Fully Differentiable Architecture Search}} for
  {{Generative Adversarial Networks}}.
\newblock \emph{arXiv:2006.09134 [cs, eess]}, 2020{\natexlab{b}}.

\bibitem[Tolstikhin et~al.(2017)Tolstikhin, Gelly, Bousquet, {Simon-Gabriel},
  and Sch{\"o}lkopf]{tolstikhin2017AdaGAN}
Ilya Tolstikhin, Sylvain Gelly, Olivier Bousquet, Carl-Johann {Simon-Gabriel},
  and Bernhard Sch{\"o}lkopf.
\newblock {{AdaGAN}}: {{Boosting Generative Models}}.
\newblock \emph{arXiv:1701.02386 [cs, stat]}, 2017.

\bibitem[Wang \& Huan(2019)Wang and Huan]{wang2019AGAN}
Hanchao Wang and Jun Huan.
\newblock {{AGAN}}: {{Towards Automated Design}} of {{Generative Adversarial
  Networks}}.
\newblock \emph{arXiv:1906.11080 [cs, stat]}, 2019.

\bibitem[Williams(1992)]{williams1992Simple}
Ronald~J. Williams.
\newblock Simple statistical gradient-following algorithms for connectionist
  reinforcement learning.
\newblock \emph{Machine Learning}, 8\penalty0 (3), 1992.
\newblock ISSN 1573-0565.
\newblock \doi{10.1007/BF00992696}.

\bibitem[Xu et~al.(2020)Xu, Xie, Zhang, Chen, Qi, Tian, and Xiong]{xu2020pc}
Yuhui Xu, Lingxi Xie, Xiaopeng Zhang, Xin Chen, Guo-Jun Qi, Qi~Tian, and
  Hongkai Xiong.
\newblock Pc-darts: Partial channel connections for memory-efficient
  architecture search.
\newblock In \emph{ICLR}, 2020.

\bibitem[Yang et~al.(2020)Yang, Shen, and Zhou]{yang2020Semantic}
Ceyuan Yang, Yujun Shen, and Bolei Zhou.
\newblock Semantic {{Hierarchy Emerges}} in {{Deep Generative Representations}}
  for {{Scene Synthesis}}.
\newblock \emph{arXiv:1911.09267 [cs]}, 2020.

\bibitem[Ying et~al.(2019)Ying, Klein, Real, Christiansen, Murphy, and
  Hutter]{ying2019NASBench101}
Chris Ying, Aaron Klein, Esteban Real, Eric Christiansen, Kevin Murphy, and
  Frank Hutter.
\newblock {{NAS}}-{{Bench}}-101: {{Towards Reproducible Neural Architecture
  Search}}.
\newblock \emph{arXiv:1902.09635 [cs, stat]}, 2019.

\bibitem[Zhang et~al.(2018)Zhang, Goodfellow, Metaxas, and
  Odena]{zhang2018SelfAttention}
Han Zhang, Ian Goodfellow, Dimitris Metaxas, and Augustus Odena.
\newblock Self-{{Attention Generative Adversarial Networks}}.
\newblock \emph{arXiv:1805.08318 [cs, stat]}, 2018.

\bibitem[Zhao et~al.(2020)Zhao, Li, Yu, Gao, and Chen]{zhao2020Feature}
Yang Zhao, Chunyuan Li, Ping Yu, Jianfeng Gao, and Changyou Chen.
\newblock Feature {{Quantization Improves GAN Training}}.
\newblock \emph{arXiv:2004.02088 [cs, stat]}, 2020.

\end{thebibliography}


\begin{thebibliography}{0}
\providecommand{\natexlab}[1]{#1}
\providecommand{\url}[1]{\texttt{#1}}
\expandafter\ifx\csname urlstyle\endcsname\relax
  \providecommand{\doi}[1]{doi: #1}\else
  \providecommand{\doi}{doi: \begingroup \urlstyle{rm}\Url}\fi

\end{thebibliography}
\bibliographystyle{iclr2021_conference}

\appendix

\clearpage
\section{Search Algorithm}
\label{apx-sec:search_algorithm}

Algorithm~\ref{alg:algorithm} shows the pseudo-code of the search procedure.
As shown on line 2, we employ fair sampling strategy~\citep{chu2019FairNAS} to offer equal opportunity for training each child model of the super-network. However, unlike FairNAS~\citep{chu2019FairNAS} performing architecture search after the training of the super-network, our method embed the search process (policy learning) into the training loop of the super-network (shown on line 12). This is equivalent to performing a moving average for the policy parameters over the course of training. Based on \textsf{NAS-caGAN} model, we investigate the difference between these two search strategies. On CIFAR10, these models report FID scores of $6.83$ and $7.61$, respectively, \ie, our proposed search strategy is better. We attribute the performance gain to the moving average for the policy parameters that could reduce the ranking noise of the weight-sharing NAS~\citep{chen2020Fitting}.

Thanks to the proposed mixed-architecture optimization, the search procedure can be very simple regardless of the number of classes. As shown on line 8, mixed-architecture optimization allows class image generation with distinct generating architectures in a single forward pass. That is to say, although each class may have varied generator architecture, these architectures can forward in parallel so that the training process is as simple as that of the original GAN~\citep{goodfellow2014Generative}.

\begin{algorithm}[!t]
   \caption{Searching with mixed-architecture optimization (pseudo-code in a PyTorch style).}
   \label{alg:algorithm}
   \begin{algorithmic}[1]
      \Statex \textcolor{Aquamarine}{\# $N_\mathrm{op}$ \hspace{0.9cm}:the number of operators in a searched edge}
      \Statex \textcolor{Aquamarine}{\# $L$ \hspace{1.25cm}:the number of searched edges in an architecture}
      \Statex \textcolor{Aquamarine}{\# $N_\mathrm{critic}$ \hspace{0.55cm}:the number of times for updating the discriminator per update of the generator}
      \Statex \textcolor{Aquamarine}{\# $N_\mathrm{policy}$ \hspace{0.44cm}:update policy every $N_\mathrm{policy}$ iterations}
      \Statex \textcolor{Aquamarine}{\# $N_\mathrm{c}$ \hspace{0.98cm} :the number of classes}
      \Statex

      \State for \ $\mathrm{iter}, (\boldsymbol{x}, y)$ in enumerate(data\_loader):
      \quad\textcolor{Aquamarine}{\# images $\boldsymbol{x}: (b, c, h, w)$, class label $y: (b, )$}
      \Statex \quad\textcolor{Aquamarine}{\# prepare a batch of network architectures}
      \\ \quad $\mathrm{fair\_arcs}=\mathrm{fair\_arc\_sampling}()$
      \quad\textcolor{Aquamarine}{\# $\mathrm{fair\_arcs}: (N_\mathrm{op}, L)$, derived by the fair sampling}
      \\ \quad $\mathrm{arcs}=\mathrm{fair\_arcs}.\mathrm{repeat}(b, 1)$
      \quad\textcolor{Aquamarine}{\# $\mathrm{arcs}: (b\times N_\mathrm{op}, L)$}
      \Statex

      \Statex \quad\textcolor{Aquamarine}{\# broadcast inputs to ensure each sample goes through $fair\_arcs$}
      \\ \quad $\boldsymbol{x}=\boldsymbol{x}.\mathrm{repeat\_interleave}(\mathrm{repeats}=N_\mathrm{op}, \mathrm{dim}=0)$
      \quad\textcolor{Aquamarine}{\# $\boldsymbol{x}:(b\times N_\mathrm{op}, c, h, w)$}
      \\ \quad $y=y.\mathrm{repeat\_interleave}(\mathrm{repeats}=N_\mathrm{op}, \mathrm{dim}=0)$
      \quad\textcolor{Aquamarine}{\# $y:(b\times N_\mathrm{op}, )$}
      \Statex

      \Statex \quad\textcolor{Aquamarine}{\# update super-network parameters $\omega$ (we simplify the notation by omitting the noise $\boldsymbol{z}$)}
      \\ \quad update\_D$(\boldsymbol{x}, y, \mathrm{arcs})$
      \\ \quad if \ $\mathrm{iter} \ \mathrm{\%} \ N_\mathrm{critic} == 0$:
      \\ \quad\quad update\_G$(y, \mathrm{arcs})$
      \Statex

      \Statex \quad\textcolor{Aquamarine}{\# update policy parameters $\boldsymbol{\theta}$}
      \\ \quad if \ $\mathrm{iter} \ \mathrm{\%} \ N_\mathrm{policy} == 0$:
      \Statex \quad\quad\textcolor{Aquamarine}{\# sample generator architectures by the policy for all classes}
      \\ \quad\quad $\mathrm{class\_arcs}=\mathrm{arc\_sampling\_by\_policy()}$
      \quad\textcolor{Aquamarine}{\# $\mathrm{class\_arcs}: (N_\mathrm{c}, L)$}
      \\ \quad\quad $\mathrm{reward}=\mathrm{eval\_InceptionScore}(\mathrm{class\_arcs})$
      \\ \quad\quad update\_policy$(\mathrm{reward})$

      \Statex
      \vspace{0.2cm}
      {\bf Return:}
      policy $\pi(a;\boldsymbol{\theta})$
   \end{algorithmic}
\end{algorithm}

\section{Implementation Details}
\label{apx:details}

We use the Adam optimizer~\citep{kingma2014Adam} with ${\eta}={0.0001}$, ${\beta_1}={0}$, and ${\beta_2}={0.9}$ for optimizing GAN, and ${\eta}={0.00035}$, ${\beta_1}={0.9}$, and ${\beta_2}={0.999}$ for policy learning. For the discriminator, we adopt the same architecture used in AutoGAN~\citep{gong2019AutoGAN}, but equip it with the \textit{cproj}~\citep{miyato2018cGANs}. The discriminator is updated five times per update of the generator. All experiments are performed on a GeForce GTX-1080Ti GPU, with the batch size set to be $32$. We search for $200\mathrm{K}$ iterations and re-train for $500\mathrm{K}$ iterations that are sufficient to achieve stable results.

\section{Searched Generator Architectures on CIFAR-10}
\label{apx-sec:class_aware_arcs_cifar10}

\subsection{Generator architectures for \textsf{NAS-cGAN} and \textsf{NAS-caGAN}}
\label{apx-sec:arcs-nas-cgan-nas-cagan}

We use \textit{RConv}$\_3\times3$ and \textit{CMConv}$\_3\times3$ as candidate operators. These operators and their corresponding index numbers are shown in Table~\ref{tab:nas-cgan-nas-cagan-cifar10}. On CIFAR10, there are three cells for a generator architecture, each of which contains two nodes, so there are three edges to be searched in a cell and nine edges in total. The searched generator architectures of \textsf{NAS-cGAN} and \textsf{NAS-caGAN} are shown in Table~\ref{tab:arcs-nas-cgan-nas-cagan-cifar10}. It can be seen that the class conditional operator (\textit{CMConv}) tends to appear at the shallow layers (close to the input noise vector), and the class unrelated operator (\textit{RConv}) prefers to appear at the output layers (close to the generated images). This phenomenon is definitely suggestive for future studies on the design of generator architectures.

\begin{table}[H]
   \centering
   \caption{Candidate operators for \textsf{NAS-cGAN} and \textsf{NAS-caGAN}}
   \begin{tabular}{c|l}
      \toprule
      Index & \multicolumn{1}{c}{Operator}                              \\ \midrule
      0     & \textit{RConv}$\_3\times3$ (regular convolution)          \\
      1     & \textit{CMConv}$\_3\times3$ (class-modulated convolution) \\
      \bottomrule
   \end{tabular}
   \label{tab:nas-cgan-nas-cagan-cifar10}
\end{table}

\begin{table}[H]
   \centering
   \caption{Searched architectures for \textsf{NAS-cGAN} and \textsf{NAS-caGAN}}
   \label{tab:arcs-nas-cgan-nas-cagan-cifar10}
   \begin{tabular}{l|c|ccccccccc}
      \toprule
      \multirow{2}{*}{Method}              & \multirow{2}{*}{Class} & \multicolumn{9}{c}{Layer}                                                                                                                                                                                    \\ \cmidrule{3-11}
                                           &                        & \multicolumn{1}{c|}{0}    & \multicolumn{1}{c|}{1} & \multicolumn{1}{c|}{2} & \multicolumn{1}{c|}{3} & \multicolumn{1}{c|}{4} & \multicolumn{1}{c|}{5} & \multicolumn{1}{c|}{6} & \multicolumn{1}{c|}{7} & 8 \\
      \midrule
      \textsf{NAS-cGAN}                    & \textit{All}           & 1                         & 1                      & 0                      & 1                      & 0                      & 0                      & 0                      & 0                      & 0 \\
      \midrule
      \multirow{10}{*}{\textsf{NAS-caGAN}} & 0                      & 1                         & 1                      & 0                      & 0                      & 0                      & 0                      & 0                      & 0                      & 0 \\
                                           & 1                      & 1                         & 1                      & 1                      & 0                      & 0                      & 1                      & 0                      & 1                      & 0 \\
                                           & 2                      & 1                         & 0                      & 1                      & 0                      & 1                      & 1                      & 0                      & 0                      & 1 \\
                                           & 3                      & 1                         & 1                      & 1                      & 0                      & 0                      & 1                      & 1                      & 1                      & 0 \\
                                           & 4                      & 1                         & 0                      & 1                      & 0                      & 1                      & 0                      & 0                      & 1                      & 0 \\
                                           & 5                      & 1                         & 0                      & 1                      & 1                      & 1                      & 1                      & 1                      & 1                      & 0 \\
                                           & 6                      & 1                         & 0                      & 1                      & 0                      & 1                      & 0                      & 1                      & 0                      & 0 \\
                                           & 7                      & 1                         & 0                      & 1                      & 1                      & 1                      & 0                      & 0                      & 1                      & 0 \\
                                           & 8                      & 1                         & 1                      & 1                      & 1                      & 1                      & 0                      & 1                      & 0                      & 0 \\
                                           & 9                      & 1                         & 1                      & 1                      & 0                      & 0                      & 0                      & 0                      & 0                      & 0 \\
      \bottomrule
   \end{tabular}
\end{table}

\subsection{Comparison with state-of-the-art cGANs}
\label{apx:sec:comparison_cifar10}

Table~\ref{apx:tab:cifar10} shows some current state-of-the-art results of cGANs. We emphasize that although our results are not the best, other methods are orthogonal to ours, such as FQ-GAN~\citep{zhao2020Feature} and ADA~\citep{karras2020Training}. Combined with other approaches (\eg, FQ-GAN, ADA), our method has the potential to achieve better results.

\begin{table}[!h]
   \centering
   \caption{FID and IS scores on CIFAR10. $^\dagger$ indicates quoted from the paper.}
   \begin{tabular}{lcc}
      \toprule
      Method                                 & FID $\downarrow$        & IS $\uparrow$            \\
      \midrule
      SN-GAN~\citep{miyato2018Spectral}      & $15.73$                 & $8.19$                   \\
      \textit{cproj}~\citep{miyato2018cGANs} & $17.50^\dagger$         & $8.62^\dagger$           \\
      Multi-hinge~\citep{kavalerov2019cGANs} & $6.22$                  & $9.55$                   \\
      FQ-GAN~\citep{zhao2020Feature}         & $6.54$                  & $9.18$                   \\
      ADA~\citep{karras2020Training}         & $\textbf{2.67}^\dagger$ & $\textbf{10.06}^\dagger$ \\
      \midrule
      \textsf{NAS-cGAN} (ours)               & $6.63$                  & $9.00$                   \\
      \textsf{NAS-caGAN} (ours)              & $5.85$                  & $9.07$                   \\
      \bottomrule
   \end{tabular}
   \label{apx:tab:cifar10}
\end{table}

\subsection{Generator architectures for \textsf{NAS-caGAN-light}}

We add a non-parameter operator, \textit{zero}, to the search space, and derive a light-weighted version of class-aware generators, named \textsf{NAS-caGAN-light}. Table~\ref{tab:nas-cagan-light-cifar10} and \ref{tab:arcs-nas-cagan-light-cifar10} present the operators with index numbers and searched architectures, respectively. As shown in Table~\ref{tab:arcs-nas-cagan-light-cifar10}, the distribution of \textit{RConv} and \textit{CMConv} still conforms to the rules mentioned in Sec.~\ref{apx-sec:arcs-nas-cgan-nas-cagan}.

\begin{figure}[H]
   \centering
   \includegraphics[width=0.7\linewidth]{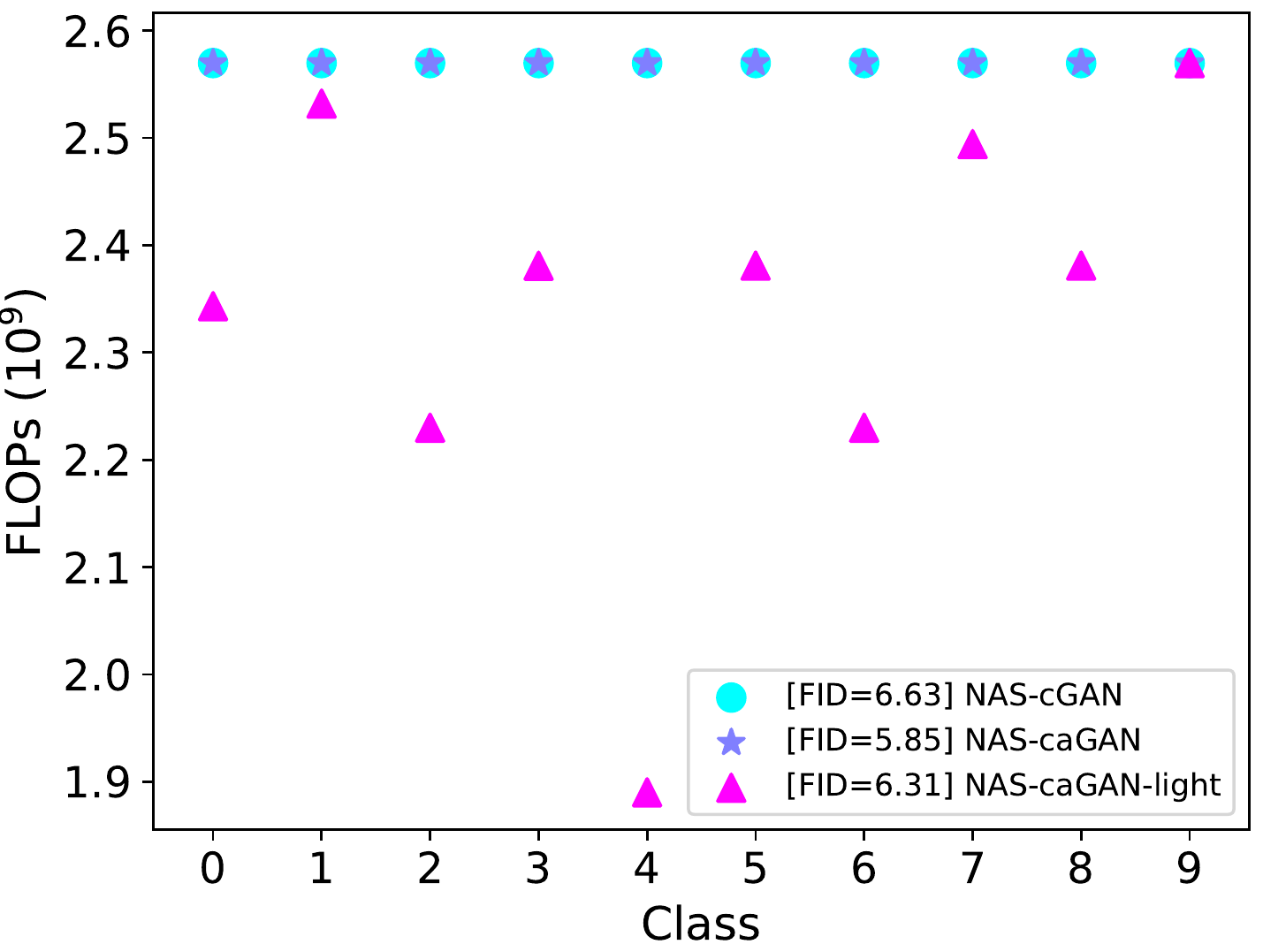}
   \caption{The computational overhead of each class on CIFAR10. Under comparable FLOPs, \textsf{NAS-caGAN} is superior to \textsf{NAS-cGAN} in terms of FID score. Under comparable FID score, \textsf{NAS-caGAN-light} has fewer FLOPs than \textsf{NAS-cGAN} for each class.}
   \label{apx:fig:flops}
\end{figure}

\begin{figure}[!th]
   \begin{subfigure}{\textwidth}
      \includegraphics[width=0.8\linewidth, left]{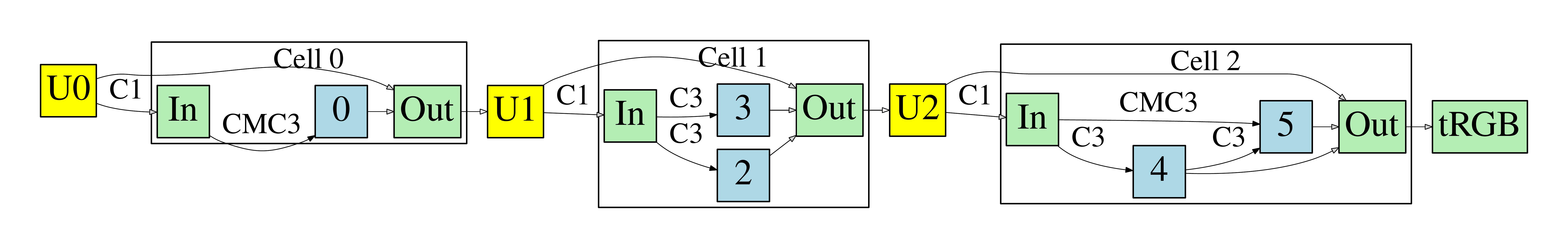}
      \vspace{-0.8cm}
      \caption{Airplane}
   \end{subfigure}
   \begin{subfigure}{\textwidth}
      \includegraphics[width=\linewidth, left]{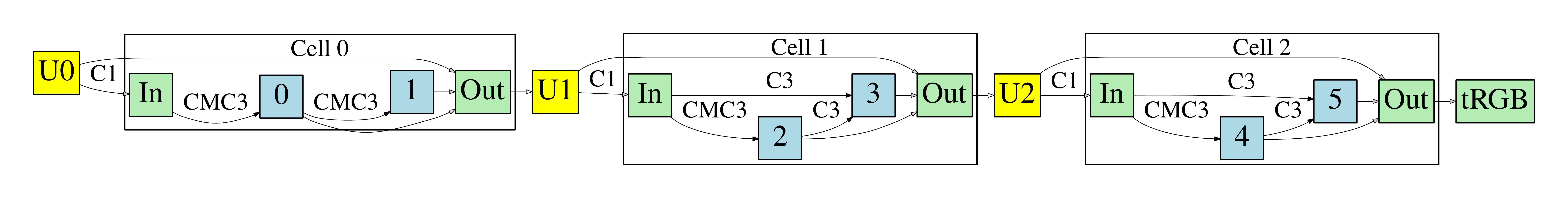}
      \vspace{-0.8cm}
      \caption{Automobile}
   \end{subfigure}
   \begin{subfigure}{\textwidth}
      \includegraphics[width=0.8\linewidth, left]{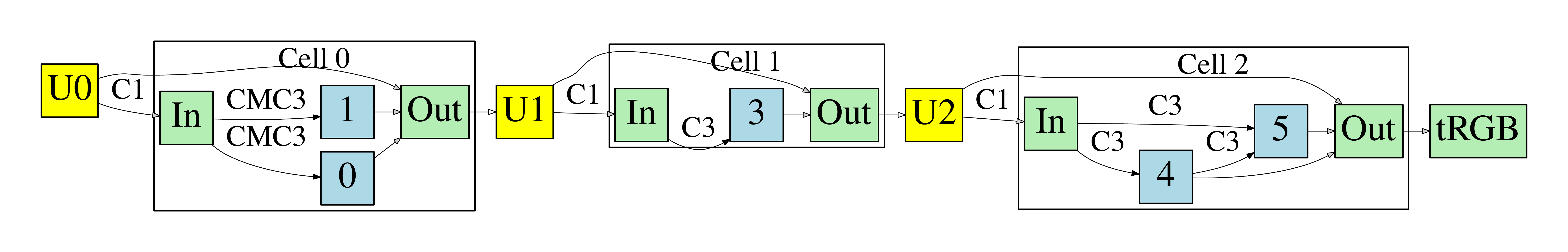}
      \vspace{-0.8cm}
      \caption{Bird}
   \end{subfigure}
   \begin{subfigure}{\textwidth}
      \includegraphics[width=0.9\linewidth, left]{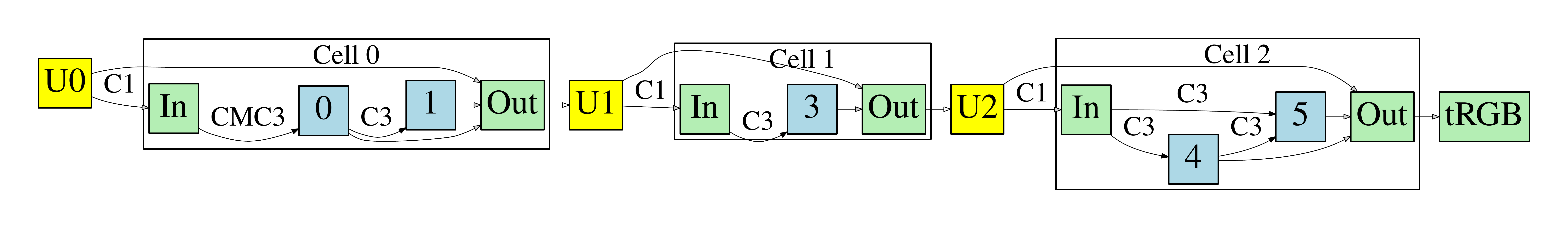}
      \vspace{-0.8cm}
      \caption{Cat}
   \end{subfigure}
   \begin{subfigure}{\textwidth}
      \includegraphics[width=0.8\linewidth, left]{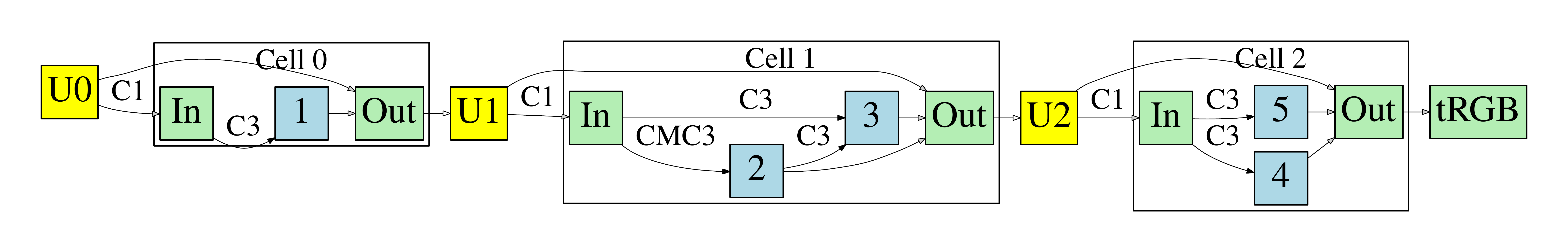}
      \vspace{-0.8cm}
      \caption{Deer}
   \end{subfigure}
   \begin{subfigure}{\textwidth}
      \includegraphics[width=0.9\linewidth, left]{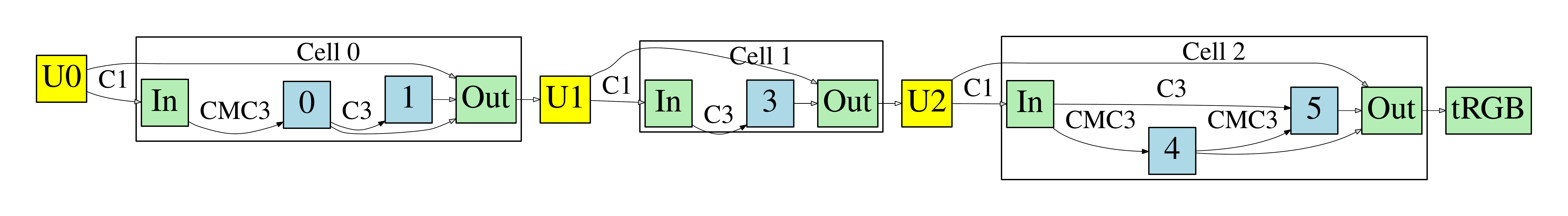}
      \vspace{-0.8cm}
      \caption{Dog}
   \end{subfigure}
   \begin{subfigure}{\textwidth}
      \includegraphics[width=0.8\linewidth, left]{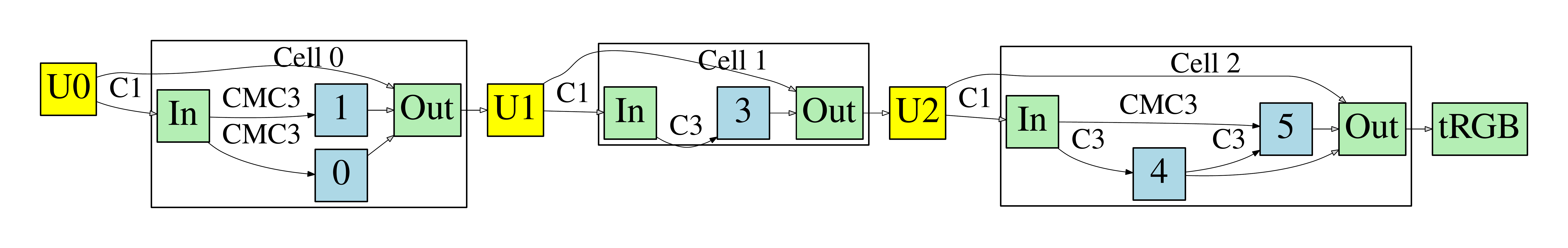}
      \vspace{-0.8cm}
      \caption{Frog}
   \end{subfigure}
   \begin{subfigure}{\textwidth}
      \includegraphics[width=0.9\linewidth, left]{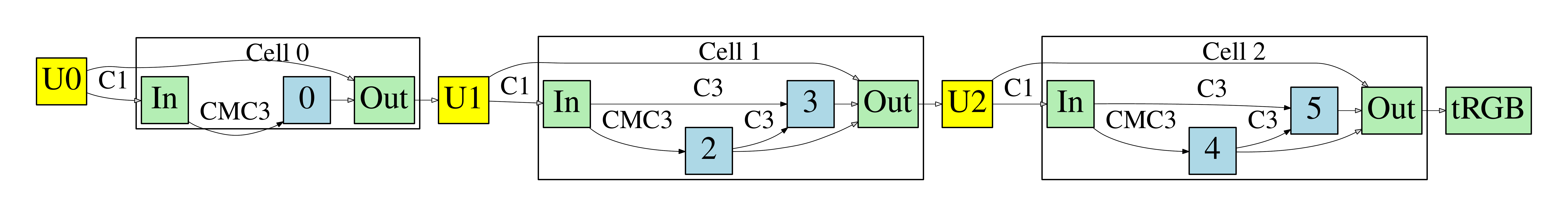}
      \vspace{-0.8cm}
      \caption{Horse}
   \end{subfigure}
   \begin{subfigure}{\textwidth}
      \includegraphics[width=0.9\linewidth, left]{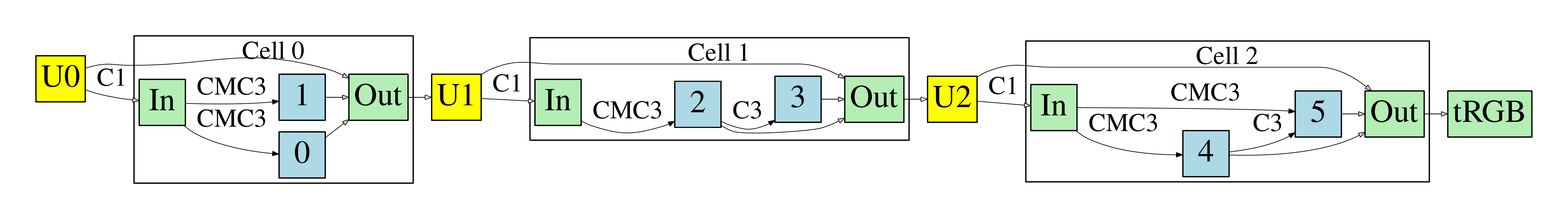}
      \vspace{-0.8cm}
      \caption{Ship}
   \end{subfigure}
   \begin{subfigure}{\textwidth}
      \includegraphics[width=\linewidth, left]{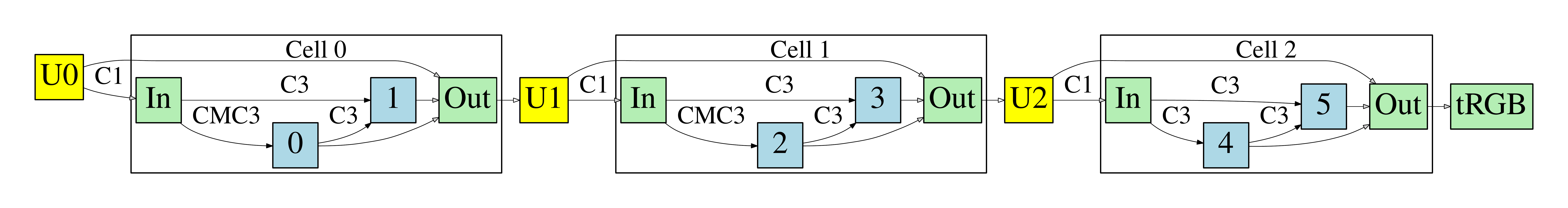}
      \vspace{-0.8cm}
      \caption{Truck}
   \end{subfigure}

   \caption{Visualization of class-aware generators of \textsf{NAS-caGAN-light}}
   \label{apx:fig:visualization}

\end{figure}

Class-aware generator architectures enjoy another merit that different classes could have varied computational overhead. Figure~\ref{apx:fig:flops} shows the computational overhead of each class on CIFAR10. It can be seen that \textsf{NAS-caGAN} is superior to \textsf{NAS-cGAN} in terms of FID score. Both models have almost the same computational overhead for all classes because all the candidate operators are parameterized (\textit{RConv} and \textit{CMConv}). After incorporating the non-parameter operator, \textsf{NAS-caGAN-light} achieves comparable FID score with \textsf{NAS-cGAN}, but with less computational overhead. The visualization of the class-aware generators of \textsf{NAS-caGAN-light} is presented in Figure~\ref{apx:fig:visualization}.

\begin{table}[H]
   \centering
   \caption{Candidate operators for \textsf{NAS-caGAN-light}}
   \begin{tabular}{c|l}
      \toprule
      Index & \multicolumn{1}{c}{Operator}                              \\ \midrule
      0     & \textit{zero}                                             \\
      1     & \textit{RConv}$\_3\times3$ (regular convolution)          \\
      2     & \textit{CMConv}$\_3\times3$ (class-modulated convolution) \\
      \bottomrule
   \end{tabular}
   \label{tab:nas-cagan-light-cifar10}
\end{table}

\begin{table}[H]
   \centering
   \caption{Searched architectures for \textsf{NAS-caGAN-light}}
   \label{tab:arcs-nas-cagan-light-cifar10}
   \begin{tabular}{l|c|ccccccccc}
      \toprule
      \multirow{2}{*}{Method}                    & \multirow{2}{*}{Class} & \multicolumn{9}{c}{Layer}                                                                                                                                                                                    \\ \cmidrule{3-11}
                                                 &                        & \multicolumn{1}{c|}{0}    & \multicolumn{1}{c|}{1} & \multicolumn{1}{c|}{2} & \multicolumn{1}{c|}{3} & \multicolumn{1}{c|}{4} & \multicolumn{1}{c|}{5} & \multicolumn{1}{c|}{6} & \multicolumn{1}{c|}{7} & 8 \\
      \midrule
      \multirow{10}{*}{\textsf{NAS-caGAN-light}} & 0                      & 2                         & 0                      & 0                      & 1                      & 1                      & 0                      & 1                      & 2                      & 1 \\
                                                 & 1                      & 2                         & 0                      & 2                      & 2                      & 1                      & 1                      & 2                      & 1                      & 1 \\
                                                 & 2                      & 2                         & 2                      & 0                      & 0                      & 1                      & 0                      & 1                      & 1                      & 1 \\
                                                 & 3                      & 2                         & 0                      & 1                      & 0                      & 1                      & 1                      & 1                      & 1                      & 1 \\
                                                 & 4                      & 0                         & 1                      & 0                      & 2                      & 1                      & 1                      & 1                      & 1                      & 0 \\
                                                 & 5                      & 2                         & 0                      & 1                      & 0                      & 1                      & 1                      & 2                      & 1                      & 2 \\
                                                 & 6                      & 2                         & 2                      & 0                      & 0                      & 1                      & 0                      & 1                      & 2                      & 1 \\
                                                 & 7                      & 2                         & 0                      & 0                      & 2                      & 1                      & 1                      & 2                      & 1                      & 1 \\
                                                 & 8                      & 2                         & 2                      & 0                      & 2                      & 0                      & 1                      & 2                      & 2                      & 1 \\
                                                 & 9                      & 2                         & 1                      & 1                      & 2                      & 1                      & 1                      & 1                      & 1                      & 1 \\
      \bottomrule
   \end{tabular}
\end{table}


\section{Searched Generator Architectures on CIFAR-100}
\label{apx-sec:class_aware_arcs_cifar100}

We perform experiments on CIFAR100, using the same search space and hyper-parameters as in CIFAR10, except increasing the number of classes to 100. That is to say,  there will be more policy parameters for the search of \textsf{NAS-caGAN}. We emphasize that owing to the mixed-architecture optimization, the search and retraining programs can be applied to this setting with more classes without any modification. The candidate operators and searched architectures are shown in Table~\ref{tab:nas-cgan-nas-cagan-cifar100} and Table~\ref{tab:cifar100-arcs}, respectively. One can see from Table~\ref{tab:cifar100-arcs} that the distribution of \textit{RConv} and \textit{CMConv} still conforms to the rules discussed in Sec.~\ref{apx-sec:arcs-nas-cgan-nas-cagan}.

\begin{table}[H]
   \centering
   \caption{Candidate operators for \textsf{NAS-cGAN} and \textsf{NAS-caGAN}}
   \begin{tabular}{c|l}
      \toprule
      Index & \multicolumn{1}{c}{Operator}                              \\ \midrule
      0     & \textit{RConv}$\_3\times3$ (regular convolution)          \\
      1     & \textit{CMConv}$\_3\times3$ (class-modulated convolution) \\
      \bottomrule
   \end{tabular}
   \label{tab:nas-cgan-nas-cagan-cifar100}
\end{table}

\begin{longtable}[c]{l|c|ccccccccc}
   \caption{Searched architectures for \textsf{NAS-cGAN} and \textsf{NAS-caGAN}}
   \label{tab:cifar100-arcs}                                                                                                                                                                                                                                                   \\
   \toprule
   \multirow{2}{*}{Method}             & \multirow{2}{*}{Class} & \multicolumn{9}{c}{Layer}                                                                                                                                                                                    \\ \cmidrule{3-11}
                                       &                        & \multicolumn{1}{c|}{0}    & \multicolumn{1}{c|}{1} & \multicolumn{1}{c|}{2} & \multicolumn{1}{c|}{3} & \multicolumn{1}{c|}{4} & \multicolumn{1}{c|}{5} & \multicolumn{1}{c|}{6} & \multicolumn{1}{c|}{7} & 8 \\
   \midrule
   \endfirsthead
   \endhead
   \hline
   \endfoot
   \endlastfoot
   \textsf{NAS-cGAN}                   & \textit{All}           & 1                         & 1                      & 0                      & 1                      & 0                      & 1                      & 1                      & 1                      & 0 \\
   \midrule
   \multirow{1}{*}{\textsf{NAS-caGAN}} & 0                      & 1                         & 1                      & 1                      & 1                      & 0                      & 1                      & 1                      & 0                      & 0 \\
                                       & 1                      & 1                         & 0                      & 0                      & 0                      & 0                      & 1                      & 0                      & 1                      & 1 \\
                                       & 2                      & 1                         & 1                      & 1                      & 0                      & 1                      & 0                      & 1                      & 1                      & 0 \\
                                       & 3                      & 1                         & 1                      & 1                      & 1                      & 0                      & 1                      & 1                      & 1                      & 0 \\
                                       & 4                      & 1                         & 0                      & 1                      & 1                      & 0                      & 0                      & 1                      & 1                      & 0 \\
                                       & 5                      & 1                         & 1                      & 1                      & 0                      & 0                      & 0                      & 0                      & 0                      & 0 \\
                                       & 6                      & 1                         & 1                      & 0                      & 0                      & 0                      & 0                      & 1                      & 0                      & 0 \\
                                       & 7                      & 1                         & 1                      & 0                      & 1                      & 1                      & 1                      & 1                      & 1                      & 1 \\
                                       & 8                      & 1                         & 0                      & 1                      & 0                      & 1                      & 1                      & 0                      & 0                      & 0 \\
                                       & 9                      & 1                         & 1                      & 0                      & 0                      & 0                      & 0                      & 1                      & 1                      & 1 \\
                                       & 10                     & 1                         & 1                      & 1                      & 1                      & 1                      & 1                      & 0                      & 0                      & 0 \\
                                       & 11                     & 1                         & 0                      & 1                      & 1                      & 1                      & 1                      & 0                      & 1                      & 0 \\
                                       & 12                     & 1                         & 0                      & 0                      & 0                      & 1                      & 0                      & 1                      & 1                      & 0 \\
                                       & 13                     & 1                         & 1                      & 1                      & 0                      & 1                      & 1                      & 0                      & 1                      & 0 \\
                                       & 14                     & 1                         & 0                      & 1                      & 0                      & 1                      & 0                      & 1                      & 1                      & 0 \\
                                       & 15                     & 1                         & 0                      & 0                      & 1                      & 1                      & 0                      & 1                      & 1                      & 0 \\
                                       & 16                     & 1                         & 1                      & 0                      & 1                      & 0                      & 1                      & 0                      & 1                      & 0 \\
                                       & 17                     & 1                         & 0                      & 1                      & 0                      & 1                      & 0                      & 1                      & 1                      & 0 \\
                                       & 18                     & 0                         & 0                      & 0                      & 0                      & 0                      & 1                      & 1                      & 0                      & 0 \\
                                       & 19                     & 1                         & 1                      & 1                      & 1                      & 1                      & 0                      & 0                      & 1                      & 0 \\
                                       & 20                     & 1                         & 1                      & 1                      & 0                      & 0                      & 1                      & 1                      & 1                      & 0 \\
                                       & 21                     & 1                         & 0                      & 1                      & 1                      & 1                      & 1                      & 1                      & 1                      & 0 \\
                                       & 22                     & 1                         & 0                      & 1                      & 0                      & 0                      & 1                      & 0                      & 1                      & 1 \\
                                       & 23                     & 1                         & 0                      & 1                      & 1                      & 1                      & 1                      & 1                      & 1                      & 1 \\
                                       & 24                     & 1                         & 0                      & 1                      & 1                      & 1                      & 0                      & 1                      & 0                      & 0 \\
                                       & 25                     & 1                         & 0                      & 1                      & 0                      & 0                      & 0                      & 0                      & 1                      & 0 \\
                                       & 26                     & 1                         & 1                      & 1                      & 1                      & 0                      & 0                      & 1                      & 1                      & 0 \\
                                       & 27                     & 1                         & 1                      & 1                      & 0                      & 1                      & 0                      & 1                      & 1                      & 0 \\
                                       & 28                     & 1                         & 1                      & 1                      & 0                      & 0                      & 0                      & 1                      & 1                      & 0 \\
                                       & 29                     & 1                         & 1                      & 0                      & 0                      & 1                      & 0                      & 0                      & 0                      & 0 \\
                                       & 30                     & 1                         & 1                      & 1                      & 1                      & 1                      & 0                      & 1                      & 1                      & 0 \\
                                       & 31                     & 1                         & 0                      & 1                      & 1                      & 0                      & 0                      & 1                      & 1                      & 0 \\
                                       & 32                     & 0                         & 1                      & 1                      & 0                      & 0                      & 0                      & 0                      & 0                      & 0 \\
                                       & 33                     & 1                         & 0                      & 0                      & 0                      & 1                      & 0                      & 0                      & 1                      & 1 \\
                                       & 34                     & 1                         & 1                      & 0                      & 0                      & 0                      & 0                      & 1                      & 0                      & 0 \\
                                       & 35                     & 1                         & 1                      & 1                      & 1                      & 1                      & 0                      & 0                      & 1                      & 0 \\
                                       & 36                     & 0                         & 1                      & 1                      & 1                      & 0                      & 1                      & 0                      & 0                      & 1 \\
                                       & 37                     & 1                         & 0                      & 1                      & 0                      & 1                      & 0                      & 0                      & 0                      & 0 \\
                                       & 38                     & 1                         & 1                      & 1                      & 1                      & 0                      & 0                      & 1                      & 1                      & 0 \\
                                       & 39                     & 1                         & 0                      & 1                      & 0                      & 1                      & 1                      & 1                      & 1                      & 0 \\
                                       & 40                     & 1                         & 1                      & 1                      & 0                      & 1                      & 0                      & 1                      & 0                      & 0 \\
                                       & 41                     & 1                         & 0                      & 1                      & 0                      & 1                      & 0                      & 0                      & 0                      & 0 \\
                                       & 42                     & 1                         & 0                      & 1                      & 0                      & 1                      & 0                      & 1                      & 0                      & 0 \\
                                       & 43                     & 1                         & 1                      & 1                      & 1                      & 1                      & 1                      & 1                      & 1                      & 0 \\
                                       & 44                     & 1                         & 0                      & 0                      & 1                      & 1                      & 0                      & 1                      & 1                      & 0 \\
                                       & 45                     & 0                         & 1                      & 0                      & 0                      & 1                      & 1                      & 0                      & 1                      & 0 \\
                                       & 46                     & 1                         & 1                      & 1                      & 1                      & 1                      & 0                      & 0                      & 0                      & 0 \\
                                       & 47                     & 0                         & 0                      & 1                      & 0                      & 0                      & 0                      & 0                      & 1                      & 0 \\
                                       & 48                     & 1                         & 1                      & 1                      & 0                      & 1                      & 1                      & 0                      & 1                      & 0 \\
                                       & 49                     & 1                         & 1                      & 0                      & 1                      & 1                      & 1                      & 1                      & 1                      & 1 \\
                                       & 50                     & 1                         & 1                      & 1                      & 1                      & 1                      & 0                      & 1                      & 1                      & 0 \\
                                       & 51                     & 1                         & 0                      & 0                      & 0                      & 1                      & 1                      & 1                      & 1                      & 0 \\
                                       & 52                     & 1                         & 1                      & 0                      & 0                      & 0                      & 0                      & 1                      & 1                      & 0 \\
                                       & 53                     & 1                         & 1                      & 1                      & 1                      & 1                      & 1                      & 1                      & 0                      & 1 \\
                                       & 54                     & 1                         & 0                      & 1                      & 0                      & 0                      & 0                      & 0                      & 1                      & 0 \\
                                       & 55                     & 1                         & 0                      & 0                      & 0                      & 1                      & 1                      & 0                      & 0                      & 0 \\
                                       & 56                     & 1                         & 1                      & 1                      & 0                      & 0                      & 0                      & 0                      & 0                      & 0 \\
                                       & 57                     & 1                         & 1                      & 1                      & 0                      & 0                      & 1                      & 0                      & 0                      & 0 \\
                                       & 58                     & 1                         & 1                      & 0                      & 0                      & 0                      & 0                      & 0                      & 1                      & 0 \\
                                       & 59                     & 1                         & 1                      & 1                      & 0                      & 0                      & 0                      & 1                      & 1                      & 0 \\
                                       & 60                     & 1                         & 1                      & 1                      & 0                      & 1                      & 0                      & 1                      & 1                      & 0 \\
                                       & 61                     & 1                         & 0                      & 1                      & 0                      & 1                      & 0                      & 0                      & 0                      & 0 \\
                                       & 62                     & 0                         & 0                      & 1                      & 0                      & 0                      & 1                      & 0                      & 0                      & 1 \\
                                       & 63                     & 0                         & 0                      & 0                      & 1                      & 0                      & 0                      & 1                      & 1                      & 0 \\
                                       & 64                     & 1                         & 0                      & 1                      & 1                      & 1                      & 1                      & 1                      & 0                      & 1 \\
                                       & 65                     & 0                         & 0                      & 1                      & 1                      & 0                      & 0                      & 1                      & 0                      & 0 \\
                                       & 66                     & 1                         & 0                      & 1                      & 0                      & 0                      & 1                      & 1                      & 1                      & 0 \\
                                       & 67                     & 1                         & 0                      & 1                      & 1                      & 0                      & 1                      & 1                      & 1                      & 1 \\
                                       & 68                     & 0                         & 1                      & 1                      & 0                      & 0                      & 0                      & 1                      & 0                      & 0 \\
                                       & 69                     & 1                         & 1                      & 1                      & 0                      & 1                      & 0                      & 1                      & 1                      & 0 \\
                                       & 70                     & 1                         & 0                      & 1                      & 1                      & 0                      & 0                      & 0                      & 0                      & 1 \\
                                       & 71                     & 1                         & 1                      & 1                      & 1                      & 1                      & 1                      & 1                      & 1                      & 0 \\
                                       & 72                     & 1                         & 0                      & 0                      & 1                      & 0                      & 1                      & 1                      & 1                      & 1 \\
                                       & 73                     & 1                         & 1                      & 0                      & 1                      & 0                      & 1                      & 1                      & 1                      & 1 \\
                                       & 74                     & 0                         & 1                      & 1                      & 1                      & 0                      & 0                      & 1                      & 1                      & 0 \\
                                       & 75                     & 1                         & 0                      & 1                      & 1                      & 0                      & 1                      & 1                      & 1                      & 0 \\
                                       & 76                     & 1                         & 0                      & 1                      & 0                      & 1                      & 1                      & 0                      & 1                      & 0 \\
                                       & 77                     & 1                         & 0                      & 0                      & 1                      & 1                      & 0                      & 1                      & 0                      & 0 \\
                                       & 78                     & 1                         & 0                      & 1                      & 0                      & 1                      & 0                      & 0                      & 1                      & 0 \\
                                       & 79                     & 1                         & 1                      & 1                      & 0                      & 1                      & 1                      & 0                      & 0                      & 0 \\
                                       & 80                     & 1                         & 1                      & 1                      & 0                      & 0                      & 0                      & 1                      & 0                      & 0 \\
                                       & 81                     & 1                         & 1                      & 0                      & 0                      & 1                      & 0                      & 0                      & 0                      & 0 \\
                                       & 82                     & 0                         & 1                      & 1                      & 0                      & 0                      & 1                      & 1                      & 1                      & 0 \\
                                       & 83                     & 1                         & 0                      & 1                      & 0                      & 0                      & 1                      & 0                      & 0                      & 0 \\
                                       & 84                     & 1                         & 0                      & 0                      & 0                      & 0                      & 0                      & 0                      & 0                      & 0 \\
                                       & 85                     & 1                         & 1                      & 0                      & 0                      & 1                      & 0                      & 0                      & 1                      & 0 \\
                                       & 86                     & 1                         & 0                      & 0                      & 0                      & 1                      & 0                      & 0                      & 1                      & 0 \\
                                       & 87                     & 1                         & 1                      & 1                      & 0                      & 0                      & 1                      & 0                      & 0                      & 1 \\
                                       & 88                     & 1                         & 0                      & 1                      & 0                      & 0                      & 0                      & 1                      & 1                      & 0 \\
                                       & 89                     & 1                         & 1                      & 1                      & 0                      & 1                      & 0                      & 0                      & 1                      & 0 \\
                                       & 90                     & 1                         & 1                      & 0                      & 0                      & 1                      & 0                      & 0                      & 1                      & 0 \\
                                       & 91                     & 0                         & 1                      & 1                      & 0                      & 1                      & 1                      & 1                      & 0                      & 1 \\
                                       & 92                     & 0                         & 1                      & 1                      & 0                      & 1                      & 1                      & 1                      & 1                      & 1 \\
                                       & 93                     & 1                         & 1                      & 1                      & 0                      & 0                      & 1                      & 0                      & 0                      & 1 \\
                                       & 94                     & 1                         & 0                      & 1                      & 0                      & 0                      & 0                      & 0                      & 1                      & 0 \\
                                       & 95                     & 1                         & 1                      & 0                      & 1                      & 1                      & 1                      & 1                      & 1                      & 0 \\
                                       & 96                     & 0                         & 1                      & 0                      & 0                      & 1                      & 0                      & 0                      & 0                      & 0 \\
                                       & 97                     & 1                         & 1                      & 1                      & 0                      & 1                      & 0                      & 1                      & 0                      & 0 \\
                                       & 98                     & 1                         & 0                      & 1                      & 1                      & 1                      & 1                      & 0                      & 0                      & 0 \\
                                       & 99                     & 0                         & 1                      & 1                      & 0                      & 0                      & 1                      & 1                      & 0                      & 0 \\ \hline
\end{longtable}

\section{Unconditional Image Generation on CIFAR10}
\label{apx:sec:unconditional}

\begin{wraptable}{r}{8cm}
   \vspace{-0.48cm}
   \caption{Comparison between existing automatically designed GAN methods and our algorithm in the setting that all classes share the same generator architecture. No class label is used at all.}
   \label{tab:agnostic-cifar10}
   \centering
   \resizebox{\linewidth}{!}{
      \begin{tabular}{lcccc}
         \toprule
         Method                                       & Params (M) & FID $\downarrow$ & IS $\uparrow$            \\
         \midrule
         AGAN~\citep{wang2019AGAN}                    & $20.1$     & $30.50$          & $8.29 \pm 0.09$          \\
         AutoGAN~\citep{gong2019AutoGAN}              & $4.4$      & $12.42$          & $8.55 \pm 0.10$          \\
         AdversarialNAS~\citep{gao2019AdversarialNAS} & $8.8$      & $10.87$          & $\textbf{8.74} \pm 0.07$ \\
         \midrule
         baseline (full)                              & $6.0$      & $12.26$          & $8.61 \pm 0.07$          \\
         \textsf{NAS-GAN} (searched)                  & $5.4$      & $\textbf{10.80}$ & $8.32 \pm 0.09$          \\
         \bottomrule
      \end{tabular}
   }
\end{wraptable}

We study the effectiveness of the search space and the search algorithm. We design two class-agnostic versions based on the architecture used in class-aware setting. The first one is to directly set all operators to be \textit{RConv}, and the second is to search a common architecture for all classes, with each the operator on each edge chosen among \textit{RConv}, \textit{skip-connect}, and \textit{zero}. Note that we remove the \textit{CMConv} operator to guarantee that class information is NOT used at all. The discriminator is also class-agnostic with the same architecture used in AutoGAN (without \textit{cproj}).

Results are summarized in Table~\ref{tab:agnostic-cifar10}. We deliver two messages. First, our search space that contains $9$ selectable operators seems small, but has sufficient ability to find powerful models and compete against recently published methods. Second, the searched \textsf{NAS-GAN} preserves most parameterized operator (\textit{RConv}), showing that GAN needs a sufficient amount parameters to achieve good performance. This supports the design of our search space, \ie, the competition is held between two parameterized operators. The searched architecture is shown in Table~\ref{apx:tab:arcs-nas-gan}.




\begin{table}[H]
   \centering
   \caption{Candidate operators for \textsf{NAS-GAN}}
   \begin{tabular}{c|l}
      \toprule
      Index & \multicolumn{1}{c}{Operator}                     \\ \midrule
      0     & \textit{zero}                                    \\
      1     & \textit{skip-connect}                            \\
      2     & \textit{RConv}$\_3\times3$ (regular convolution) \\
      \bottomrule
   \end{tabular}
   \label{apx:tab:nas-gan-cifar10}
\end{table}

\begin{table}[H]
   \centering
   \caption{Architectures for baseline and \textsf{NAS-GAN}. The candidate operators and their corresponding indices are shown in Table~\ref{apx:tab:nas-gan-cifar10} }
   \label{apx:tab:arcs-nas-gan}
   \begin{tabular}{l|c|ccccccccc}
      \toprule
      \multirow{2}{*}{Method}     & \multirow{2}{*}{Class} & \multicolumn{9}{c}{Layer}                                                                                                                                                                                    \\ \cmidrule{3-11}
                                  &                        & \multicolumn{1}{c|}{0}    & \multicolumn{1}{c|}{1} & \multicolumn{1}{c|}{2} & \multicolumn{1}{c|}{3} & \multicolumn{1}{c|}{4} & \multicolumn{1}{c|}{5} & \multicolumn{1}{c|}{6} & \multicolumn{1}{c|}{7} & 8 \\
      \midrule
      baseline (full)             & \textit{All}           & 2                         & 2                      & 2                      & 2                      & 2                      & 2                      & 2                      & 2                      & 2 \\
      \textsf{NAS-GAN} (searched) & \textit{All}           & 2                         & 2                      & 1                      & 2                      & 2                      & 2                      & 2                      & 2                      & 2 \\
      \bottomrule
   \end{tabular}
\end{table}

\section{Impact of Injection Position of Class Information on Accuracy}
\label{apx:cbn_cmconv}

\begin{figure}[!t]
   \begin{subfigure}{0.49\textwidth}
      \centering
      \includegraphics[width=\linewidth]{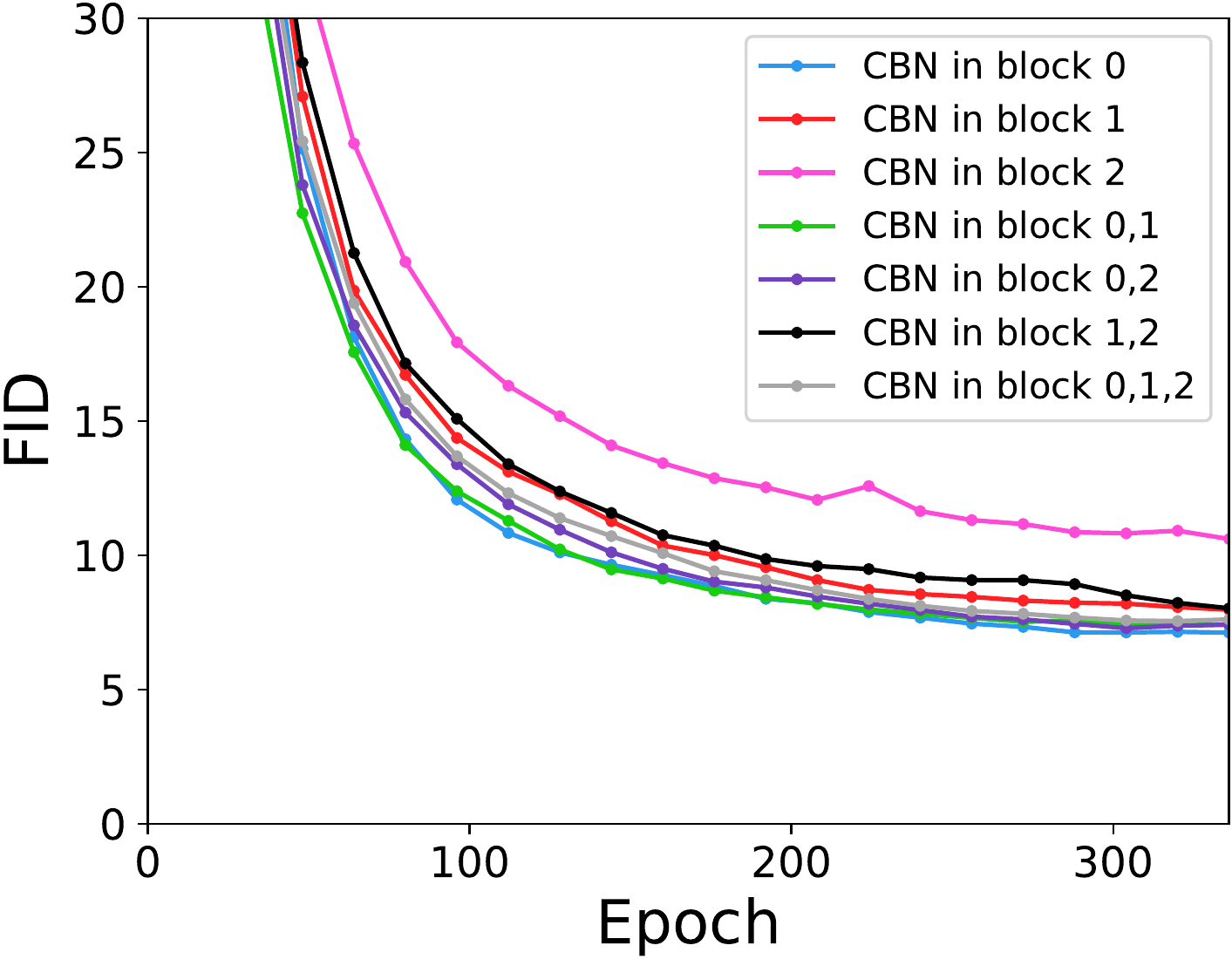}
      \caption{\textit{CBN}}
      \label{fig:sub-first}
   \end{subfigure}
   \begin{subfigure}{0.49\textwidth}
      \centering
      \includegraphics[width=\linewidth]{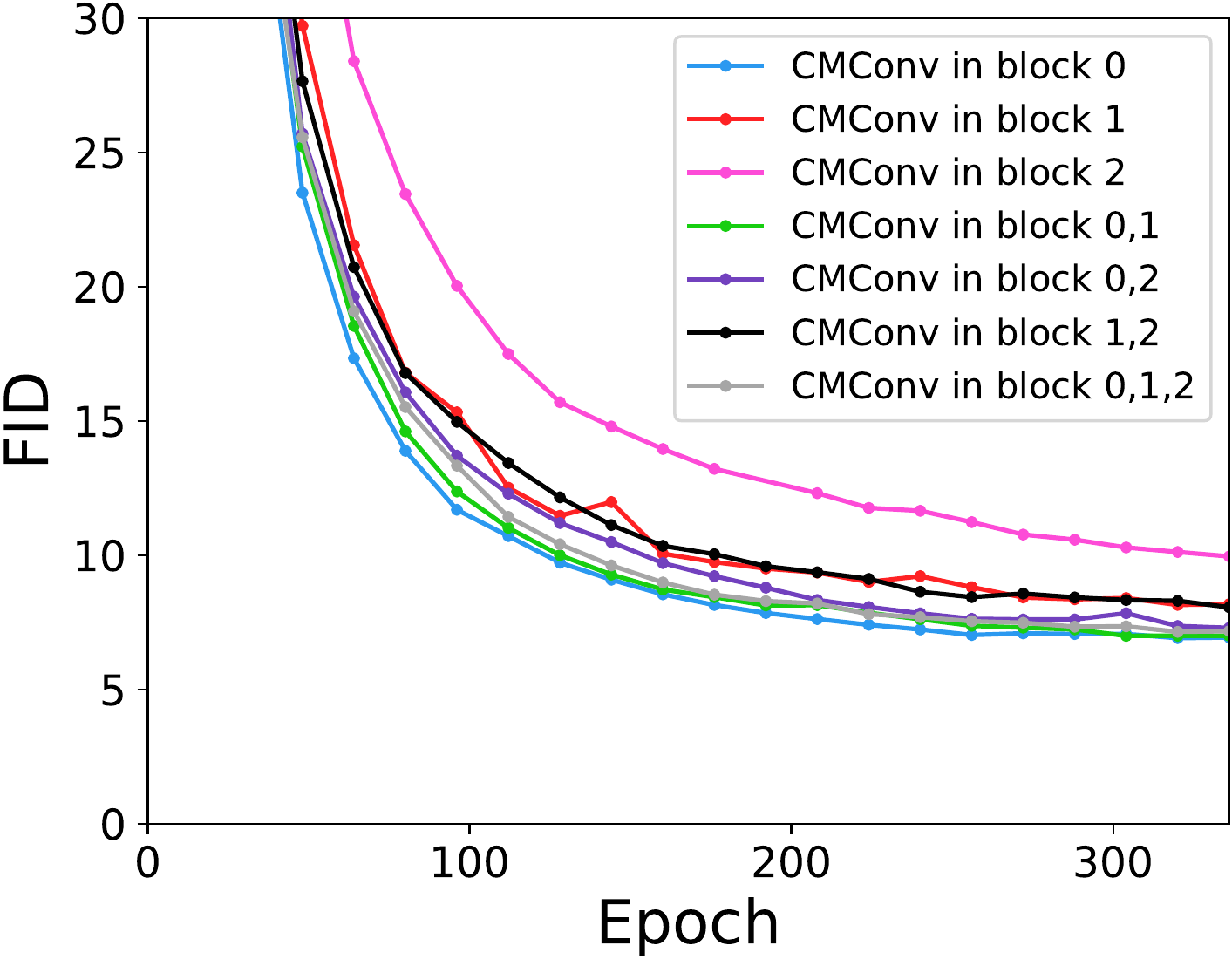}
      \caption{\textit{CMConv}}
      \label{fig:sub-first}
   \end{subfigure}
   \caption{Evaluating the effect of the inserted position of \textit{CBN} (or \textit{CMConv}) in BigGAN on CIFAR10. Inserting \textit{CBN} (\textit{CMConv}) in different positions does affect accuracy. This phenomenon is consistent with the rule we summarize through our Multi-Net NAS method.}
   \label{fig:cbn_cmconv}
\end{figure}

We use BigGAN, a manually designed GAN model, to study the impact of injection position of class information on accuracy. On CIFAR10, the generator of BigGAN has three blocks, each of which contains two class-conditional operations (\ie, \textit{CBN}). We do controlled experiments by changing the inserted position of \textit{CBN} (blocks that do not use \textit{CBN} use regular \textit{BN} as a substitute). On the other hand, we replace the regular convolution in the blocks with \textit{CMConv}, replace all \textit{CBN} with regular \textit{BN}, and study the effect of the inserted position of \textit{CMConv} on the accuracy. As shown in Figure~\ref{fig:cbn_cmconv}, The injection position of class information does have an impact on accuracy. This phenomenon has not been paid attention by previous work. We discover this phenomenon through our Multi-Net NAS.

\section{Results of Class Conditional Image Generation}

\begin{figure}[H]
   \centering
   \includegraphics[width=0.7\linewidth]{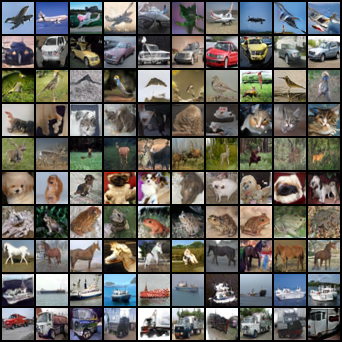}
   \caption{Generated images of \textsf{NAS-caGAN} model trained on CIFAR10. Each row corresponds to samples of the same class.}
   \label{apx:fig:c10}
\end{figure}

\begin{figure}[H]
   \centering
   \includegraphics[width=0.7\linewidth]{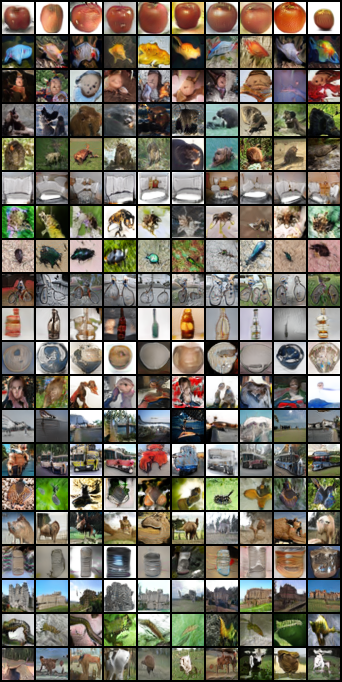}
   \caption{Generated images of \textsf{NAS-caGAN} model trained on CIFAR100. Each row corresponds to samples of the same class. Due to space limitations, we only show samples of 20 classes without cherry-picking.}
   \label{apx:fig:c100}
\end{figure}

We show some generated images of \textsf{NAS-caGAN} models trained on CIFAR10 and CIFAR100, in Figure~\ref{apx:fig:c10} and Figure~\ref{apx:fig:c100}, respectively. In the figures, each row corresponds to samples of one class. All the samples are randomly sampled rather than cherry-picked.

\end{document}